\documentclass{article}
\usepackage{ijcai24}

\usepackage{times}
\usepackage{soul}
\usepackage{url}
\usepackage[hidelinks]{hyperref}
\usepackage[utf8]{inputenc}
\usepackage[small]{caption}
\usepackage{graphicx}
\usepackage{amsmath}
\usepackage{amsthm}
\usepackage{booktabs}
\usepackage[left]{lineno}

\newtheorem{theorem}{Theorem}

\newtheorem{lemma}{Lemma}

\usepackage{amsfonts}
\usepackage{subfigure}
\usepackage{amsmath}
\usepackage{bm}
\usepackage[ruled, vlined, linesnumbered]{algorithm2e}
\usepackage{booktabs}
\usepackage{xspace}
\usepackage{color}
\makeatletter
\DeclareRobustCommand\onedot{\futurelet\@let@token\@onedot}
\def\@onedot{\ifx\@let@token.\else.\null\fi\xspace}
 
\def\ie{\emph{i.e}\onedot}

\makeatother

\title{Minimizing Weighted Counterfactual Regret with \\ Optimistic Online Mirror Descent}

\author{
    Hang Xu$^{1, 2}$\and
    Kai Li$^{1, 2,}$\thanks{Corresponding authors.}\and 
    Bingyun Liu$^{1, 2}$\and 
    Haobo Fu$^{6}$\and \\
    Qiang Fu$^{6}$\and
    Junliang Xing$^{5,}$\footnotemark[1]\And
    Jian Cheng$^{1,3,4}$\\
\affiliations
$^1$Institute of Automation, Chinese Academy of Sciences\\
$^2$School of Artificial Intelligence, University of Chinese Academy of Sciences\\
$^3$School of Future Technology, University of Chinese Academy of Sciences\\
$^4$AiRiA \\
~~~~~$^5$Tsinghua University \\
~~~~~$^6$Tencent AI Lab \\
\emails
\{xuhang2020, kai.li, liubingyun2021, jian.cheng\}@ia.ac.cn, \\
\{haobofu, leonfu\}@tencent.com, jlxing@tsinghua.edu.cn
}

\begin{document}

\maketitle

\begin{abstract}
    Counterfactual regret minimization (CFR) is a family of algorithms for effectively solving imperfect-information games.
    It decomposes the total regret into counterfactual regrets, utilizing local regret minimization algorithms, such as Regret Matching (RM) or RM+, to minimize them.
    Recent research establishes a connection between Online Mirror Descent (OMD) and RM+, paving the way for an optimistic variant PRM+ and its extension PCFR+.
    However, PCFR+ assigns uniform weights for each iteration when determining regrets, leading to substantial regrets when facing dominated actions.
    This work explores minimizing weighted counterfactual regret with optimistic OMD, resulting in a novel CFR variant PDCFR+.
    It integrates PCFR+ and Discounted CFR (DCFR) in a principled manner, swiftly mitigating negative effects of dominated actions and consistently leveraging predictions to accelerate convergence.
    Theoretical analyses prove that PDCFR+ converges to a Nash equilibrium, particularly under distinct weighting schemes for regrets and average strategies.
    Experimental results demonstrate PDCFR+'s fast convergence in common imperfect-information games.
    The code is available at \url{https://github.com/rpSebastian/PDCFRPlus}.
\end{abstract}

\section{Introduction}

Imperfect-information games (IIGs) model strategic interactions between players with hidden information.
Solving such games is challenging since players must reason under uncertainty about opponents' private information.
The hidden information plays an essential role in real-world situations such as medical treatment~\cite{sandholm2015steering}, negotiation~\cite{gratch2016misrepresentation}, and security~\cite{lisy2016counterfactual}, making the research on IIGs theoretically and practically crucial.

In this work, we focus on solving two-player zero-sum (2p0s) IIGs. The typical goal in these games is to find an (approximate) Nash equilibrium (NE) in which no player can benefit from deviating unilaterally from the equilibrium.
The common iterative approach minimizes total regrets of both players so that their average strategies over time converge to a NE.
The family of counterfactual regret minimization (CFR) algorithms~\cite{zinkevich_regret_2007} decomposes the total regret into the sum of counterfactual regrets associated with decision nodes. Then it employs a local regret minimization algorithm, such as Regret Matching (RM)~\cite{hart2000simple} or its variant RM+~\cite{tammelin_solving_2014}, at each decision node to effectively minimize counterfactual regret.
Due to the sound theoretical guarantee and strong empirical performance, CFR and its variants have enabled several major breakthroughs in this field~\cite{bowling_heads-up_2015,moravvcik2017deepstack,brown2018superhuman,brown2019superhuman}.

Besides CFR, Online Mirror Descent (OMD)~\cite{beck2003mirror} stands out as a prominent and general regret minimization algorithm.
It has promising theoretical results but remains less competitive than CFRs when directly applied to solving IIGs.
Recently, researchers have tried to build connections between CFR and OMD~\cite{farina2021faster,liu2022equivalence}.
The study by~\cite{farina2021faster} demonstrates that RM+ is essentially a specialized form of OMD when employed to minimize counterfactual regret at each decision node.
This connection inspires an optimistic variant of RM+, named Predictive RM+ (PRM+), seeking to benefit from the predictability of slowly-changing counterfactual losses over time.
Its extension Predictive CFR+ (PCFR+) exhibits extremely fast convergence on non-poker IIGs.

Despite the notable success of PCFR+, a limitation arises from its practice of assigning uniform weights to each iteration when determining regrets. The algorithm becomes particularly challenging when dealing with dominated actions, leading to high regrets for other actions and requiring a substantial number of iterations to mitigate this negative impact\footnote{We provide a motivating example in section~\ref{sec:motivation}.}.
Assigning more weights to recent iterations has shown to be important for fast convergence~\cite{tammelin_solving_2014,brown_solving_2019,grand2023solving}.
Notably, Discounted CFR (DCFR)~\cite{brown_solving_2019} is a family of algorithms that discounts prior iterations when determining both regrets and average strategy.
Thus, it is desirable to design a CFR variant that not only allocates more weights to recent iteration, thereby alleviating high regrets from earlier iterations, but also leverages predictions to accelerate convergence.

To this end, we delve into minimizing weighted counterfactual regret with OMD and optimistic variant:
\begin{itemize}
    \item We demonstrate that by directly employing OMD to minimize weighted counterfactual regret, we obtain a CFR variant, DCFR+, which incorporates regret discounting similar to DCFR and clips negative regrets akin to CFR+. Remarkably, the CFR variant previously discovered through evolutionary search~\cite{xu2022autocfr} is a special case of DCFR+, showcasing faster convergence than DCFR.
    \item Furthermore, by applying optimistic OMD, we derive a novel CFR variant, PDCFR+, which leverages predicted regrets to compute new strategy like PCFR+, while updating regrets similar to DCFR+. It swiftly mitigates negative effects of dominated actions and consistently leverages predictions to speed up convergence.
    \item We observe that the updating rule of OMD with increasing weighted loss is equivalent to OMD with an increasing learning rate or a decreasing regularization term. This observation offers a partial explanation for the superior performance of DCFR.
    \item We theoretically prove that DCFR+ and PDCFR+ converge to a NE under distinct weighting schemes for regrets and average strategies, requiring the weighting scheme for regrets to be more aggressive than that for average strategies.
    \item Finally, we conduct extensive experimental evaluations of CFR variants on commonly used IIGs in the research community. We find that PDCFR+ outperforms other CFR variants by 4-8 orders of magnitude on non-poker games and small poker games, while DCFR+ converges fastest on large-scale poker games.
\end{itemize}

\section{Preliminaries}
In this section, we present the notations for formulating IIGs, introduce typical regret minimization algorithms,
and discuss the vanilla CFR algorithm along with its variants.

\subsection{Sequential Decision Process}
\textbf{Sequential decision process}~\cite{farina2019stable} is a tree-based formalism for describing each player's decision-making in IIGs, consisting of two types of nodes: decision nodes and observation nodes.
At each \textbf{decision node} $j \in \mathcal{J}$, the player selects an \textbf{action} based on a \textbf{local strategy} $\boldsymbol{x}_j \in \Delta^{n_j}$, where $\Delta^{n_j}$ is a simplex over the action set $\mathcal{A}_j \subseteq \mathcal{A}$ of size $n_j$.
After taking the action $a$, the player proceeds to an observation node $k = \rho(j, a)$.
At each \textbf{observation node} $k \in \mathcal{K}$, the player receives a \textbf{signal} $s\in S_k$ and then reaches another decision node $j' = \rho(k, s)$.
The set $C_{j, a} = \left\{\rho({\rho(j, a), s)}: s\in S_{\rho(j, a)}\right\}$ denotes all decision nodes that are earliest reachable after taking action $a$ at $j$.
We introduce a dummy root decision node $o$ with one action to ensure a unique root.
An illustration is given in Appendix~\ref{app:sdp}.

For a local strategy set $\left\{\boldsymbol{x}_j\right\}_{j \in \mathcal{J}}$, we can construct a \textbf{sequence-form strategy} $\dot{\boldsymbol{x}}$ represented as a vector indexed over $\left\{(j, a):j \in \mathcal{J}, a\in \mathcal{A}_j\right\}$.
Each entry in the vector corresponds to a pair $(j, a)$, with the value representing the product of probabilities of all actions along the path from the root  to $(j, a)$. The sequence-form strategy space is denoted by $\mathcal{X}$.

In a 2p0s IIG, where player 1 and player 2 have sequence-from strategy space $\mathcal{X}$ and $\mathcal{Y}$, and the set of decision nodes are $\mathcal{J}_x$ and $\mathcal{J}_y$, the problem of finding a \textbf{NE} can be formulated as a bilinear saddle point problem
\begin{align*}
    \min_{\dot{\boldsymbol{x}} \in \mathcal{X}} \max_{\dot{\boldsymbol{y}}\in \mathcal{Y}} \dot{\boldsymbol{x}}^\top \boldsymbol{A}\dot{\boldsymbol{y}} =  \max_{\dot{\boldsymbol{y}}\in \mathcal{Y}} \min_{\dot{\boldsymbol{x}} \in \mathcal{X}}\dot{\boldsymbol{x}}^\top \boldsymbol{A}\dot{\boldsymbol{y}},
\end{align*}
where $\boldsymbol{A}$ is a sparse payoff matrix encoding the losses for player 1. For a strategy profile $(\dot{\boldsymbol{x}}, \dot{\boldsymbol{y}})$, let $\delta_1(\dot{\boldsymbol{x}}, \dot{\boldsymbol{y}})$ denote the incentive for player 1 to unilaterally choose another strategy: $\delta_1(\dot{\boldsymbol{x}}, \dot{\boldsymbol{y}}) = \dot{\boldsymbol{x}}^\top\boldsymbol{A}\dot{\boldsymbol{y}} - \min_{\dot{\boldsymbol{x}}'\in \mathcal{X}}\dot{\boldsymbol{x}}'^\top\boldsymbol{A}\dot{\boldsymbol{y}}$. Similarly, $\delta_2(\dot{\boldsymbol{x}}, \dot{\boldsymbol{y}}) = \max_{\dot{\boldsymbol{y}}' \in \mathcal{Y}}\dot{\boldsymbol{x}}^\top \boldsymbol{A}\dot{\boldsymbol{y}}' - \dot{\boldsymbol{x}}^\top\boldsymbol{A}\dot{\boldsymbol{y}}$. An $\boldsymbol{\epsilon}$\textbf{-NE} satisfies $\delta_i(\dot{\boldsymbol{x}}, \dot{\boldsymbol{y}}) \le \epsilon, \forall i \in \left\{1,2\right\}$. The \textbf{exploitability} of $(\dot{\boldsymbol{x}}, \dot{\boldsymbol{y}})$ measures its distance from equilibrium and is defined by $e(\dot{\boldsymbol{x}}, \dot{\boldsymbol{y}})=\sum_{i \in \left\{1,2\right\}}^{}\delta_i(\dot{\boldsymbol{x}}, \dot{\boldsymbol{y}})/2$.

\subsection{Regret Minimization}

\begin{table*}[t]

    \centering
    \resizebox{0.95\textwidth}{!}{
        \begin{tabular}{llll}
            \toprule
            \toprule
            \textbf{Algorithms}                                                                                                                              & \textbf{Cumulative Regret} $\boldsymbol{R}^t_{j} $ & \textbf{New Strategy} $\boldsymbol{x}_{j}^{t+1}$ & \textbf{Cumulative Strategy} $\boldsymbol{X}^{t}$ \\ \midrule

            CFR                                                                                                                                              &
            $\boldsymbol{R}^{t-1}_{j} + \boldsymbol{r}^t_{j}$                                                                                                &
            $\left[\boldsymbol{R}_j^t\right]^+ / \left\Vert \left[\boldsymbol{R}_j^t\right]^+ \right\Vert_1$ \qquad                                          &
            $\boldsymbol{X}^{t-1} + \dot{\boldsymbol{x}}^t$                                                                                                                                                                                                                                                              \\  \midrule

            CFR+                                                                                                                                             &
            $\left[\boldsymbol{R}^{t-1}_{j}+\boldsymbol{r}^t_{j}\right]^+$                                                                                   &
            $\boldsymbol{R}^t_{j} / \left\Vert \boldsymbol{R}^t_{j} \right\Vert_1$                                                                           &
            $\boldsymbol{X}^{t-1} + t*\dot{\boldsymbol{x}}^t$                                                                                                                                                                                                                                                            \\ \midrule

            Linear CFR                                                                                                                                       &
            $\boldsymbol{R}^{t-1}_{j}+t * \boldsymbol{r}^t_{j}$                                                                                              &
            $\left[\boldsymbol{R}_j^t\right]^+ / \left\Vert \left[\boldsymbol{R}_j^t\right]^+ \right\Vert_1$                                                 &
            $\boldsymbol{X}^{t-1} + t*\dot{\boldsymbol{x}}^t$                                                                                                                                                                                                                                                            \\     \midrule

            DCFR                                                                                                                                             &
            $\begin{aligned}
                     \boldsymbol{R}^t_{j} = \boldsymbol{R}^{t - 1}_{j} \odot \boldsymbol{d}^{t-1}_{j} + \boldsymbol{r}^t_{j}, \text{where} \\
                     \boldsymbol{d}^t_{j}[a]  = \left\{
                     \begin{array}{ll}
                        \frac{t^\alpha}{t^\alpha + 1} & \text{if}\; \boldsymbol{R}^{t}_{j}[a]> 0 \\
                        \frac{t^\beta}{t^\beta + 1}   & \text{otherwise}                         \\
                    \end{array}
                     \right.
                 \end{aligned}$                         &
            $\left[\boldsymbol{R}_j^t\right]^+ / \left\Vert \left[\boldsymbol{R}_j^t\right]^+ \right\Vert_1$                                                 &
            $\boldsymbol{X}^{t-1}  (\frac{t-1}{t})^\gamma+\dot{\boldsymbol{x}}^t$                                                                                                                                                                                                                                        \\ \midrule

            \textbf{DCFR+}                                                                                                                                   &
            $\left[\boldsymbol{R}^{t-1}_{j} \frac{(t-1)^\alpha}{(t-1)^\alpha + 1} + \boldsymbol{r}^t_{j}\right]^+$                                           &
            $\boldsymbol{R}^t_{j} / \left\Vert \boldsymbol{R}^t_{j} \right\Vert_1$                                                                           &
            $\boldsymbol{X}^{t-1} (\frac{t-1}{t})^\gamma+\dot{\boldsymbol{x}}^t$                                                                                                                                                                                                                                         \\ \midrule

            PCFR+                                                                                                                                            &
            $\left[\boldsymbol{R}^{t-1}_{j}+\boldsymbol{r}^t_{j}\right]^+$                                                                                   &
            $\begin{aligned}
                      & \tilde{\boldsymbol{R}}^{t+1}_{j} / \left\Vert \tilde{\boldsymbol{R}}^{t+1}_{j} \right\Vert_1 , \text{where} \\
                      & \tilde{\boldsymbol{R}}^{t+1}_{j}=\left[\boldsymbol{R}_{j}^{t}+\boldsymbol{v}_j^{t+1}\right]^+
                 \end{aligned}$                &
            $\boldsymbol{X}^{t-1} + t^2\dot{\boldsymbol{x}}^t$                                                                                                                                                                                                                                                           \\ \midrule

            \textbf{PDCFR+}                                                                                                                                  &
            $\left[\boldsymbol{R}^{t-1}_{j}  \frac{(t-1)^\alpha}{(t-1)^\alpha + 1} + \boldsymbol{r}^t_{j}\right]^+$                                          &
            $\begin{aligned}
                      & \tilde{\boldsymbol{R}}^{t+1}_{j} / \left\Vert \tilde{\boldsymbol{R}}^{t+1}_{j} \right\Vert_1 , \text{where}                \\
                      & \tilde{\boldsymbol{R}}^{t+1}_{j}=\left[\boldsymbol{R}_{j}^{t}\frac{t^\alpha}{t^\alpha + 1}+\boldsymbol{v}_j^{t+1}\right]^+
                 \end{aligned}$ &
            $\boldsymbol{X}^{t-1}  (\frac{t-1}{t})^\gamma+\dot{\boldsymbol{x}}^t$                                                                                                                                                                                                                                        \\ \bottomrule
        \end{tabular}}
    \caption{Comparison of existing CFR variants with our proposed \textbf{DCFR+} and \textbf{PDCFR+} ($\alpha$, $\beta$, $\gamma$ are hyperparameters).}
    \label{Alg_Com}
\end{table*}

A \textbf{regret minimization} algorithm~\cite{zinkevich2003online} repeatedly plays against an unknown environment and starts with a decision $\boldsymbol{x}^1 \in \mathcal{D} \subseteq \mathbb{R}^n$, where $\mathcal{D}$ is the decision space.
In each iteration $t$, it observes a loss $\boldsymbol{\ell}^t \in \mathbb{R}^n$ from the environment and computes its next decision $\boldsymbol{x}^{t+1}$ based on past decisions $\boldsymbol{x}^1, \dots, \boldsymbol{x}^t$ and previous losses $\boldsymbol{\ell}^1, \dots, \boldsymbol{\ell}^t$.
The objective of the algorithm is to minimize the \textbf{total regret}
\begin{align*}
    R^T = \max_{\boldsymbol{x}' \in \mathcal{D}} \sum_{t=1}^{T} \left\langle \boldsymbol{\ell}^t, \boldsymbol{x}^t - \boldsymbol{x}'\right\rangle.
\end{align*}

Online Mirror Descent (OMD)~\cite{beck2003mirror} stands out as a famous regret minimization algorithm compatible with arbitrary decision spaces. It updates the decision according to $\boldsymbol{x}^1=\operatorname{argmin}_{\boldsymbol{x}' \in \mathcal{D}}\psi(\boldsymbol{x}')$, and
\begin{align*}
    \boldsymbol{x}^{t+1} = \underset{\boldsymbol{x}' \in \mathcal{D}}{\operatorname{argmin}}\; \left\{ \left\langle \boldsymbol{\ell}^t, \boldsymbol{x}' \right\rangle + \frac{1}{\eta} \mathcal{B}_\psi(\boldsymbol{x}'\mid\mid\boldsymbol{x}^t)\right\},
\end{align*}
where $\psi: \mathcal{D} \rightarrow  \mathbb{R}$ is a regularizer, $\mathcal{B}_\psi(\boldsymbol{x}' \mid \mid \boldsymbol{x})=\psi(\boldsymbol{x}') - \psi(\boldsymbol{x}) - \left\langle \nabla \psi(\boldsymbol{x}), \boldsymbol{x}' - \boldsymbol{x}\right\rangle$ is the Bregman divergence associated with $\psi$, and $\eta>0$ is an arbitrary step size. Optimistic OMD~\cite{syrgkanis2015fast} seeks to benefit from the predictability of slowly-changing loss $\boldsymbol{\ell}^t$. In iteration $t$, it predicts the next iteration's loss $\boldsymbol{\ell}^{t+1}$ as $\boldsymbol{m}^{t+1}$ and updates the decision according to $\boldsymbol{x}^1=\boldsymbol{z}^0=\operatorname{argmin}_{\boldsymbol{x}' \in \mathcal{D}}\psi(\boldsymbol{x}')$, and
\begin{align*}
     & \boldsymbol{z}^{t} = \underset{\boldsymbol{z}' \in \mathcal{D}}{\operatorname{argmin}}\;\left\langle \boldsymbol{\ell}^t, \boldsymbol{z}'\right\rangle + \frac{1}{\eta}\mathcal{B}_\psi(\boldsymbol{z}' \mid \mid \boldsymbol{z}^{t-1}),                  \\
     & \boldsymbol{x}^{t+1} = \underset{\boldsymbol{x}' \in \mathcal{D}}{\operatorname{argmin}}\; \left\{\left\langle \boldsymbol{m}^{t+1}, \boldsymbol{x}'\right\rangle + \frac{1}{\eta}\mathcal{B}_\psi(\boldsymbol{x}' \mid \mid \boldsymbol{z}^{t})\right\}.
\end{align*}

Regret Matching (RM)~\cite{hart2000simple} and Regret Matching+ (RM+)~\cite{tammelin_solving_2014} are two regret minimization algorithms operating on the simplex $\Delta^n$.
RM typically starts with a uniform random strategy $\boldsymbol{x}^1$.
On each iteration $t$, RM calculates the \textbf{instantaneous regret} $\boldsymbol{r}^t = \left\langle \boldsymbol{x}^t, \boldsymbol{\ell}^t\right\rangle\boldsymbol{1} - \boldsymbol{\ell}^t$ and then accumulates it to obtain the \textbf{cumulative regret} $\boldsymbol{R}^t = \sum_{k=1}^{t}\boldsymbol{r}^k$.
The next strategy is proportional to the positive cumulative regret, \ie, $\boldsymbol{x}^{t+1} = \left[\boldsymbol{R}^t\right]^+ / \left\Vert \left[\boldsymbol{R}^t\right]^+ \right\Vert_1$, where $[\cdot]^+ = \max\left\{\cdot, 0\right\}$. For convenience, we define $\boldsymbol{0}/0$ as the uniform distribution. RM+ is a straightforward variant of RM. It sets any action with negative cumulative regret to zero in each iteration so that it promptly reuses an action showing promise of performing well. Formally, $\boldsymbol{R}^t = [\boldsymbol{R}^{t-1} + \boldsymbol{r}^t]^+$ and $\boldsymbol{x}^{t+1} = \boldsymbol{R}^t / \left\Vert \boldsymbol{R}^t \right\Vert_1$.

\subsection{Counterfactual Regret Minimization}
Counterfactual Regret Minimization (CFR)~\cite{zinkevich_regret_2007} is one of the most popular equilibrium-finding algorithms for solving IIGs.
It operates as a regret minimization algorithm within the sequence-form strategy space.
Given a sequence-form strategy $\dot{\boldsymbol{x}}^t \in \mathcal{X}$ and a loss $\dot{\boldsymbol{\ell}}^t = \boldsymbol{A}\dot{\boldsymbol{y}}^t$,
the key concept involves constructing a \textbf{counterfactual loss} $\boldsymbol{\ell}_j^t \in \mathbb{R}^{n_j}$ for each decision node $j\in\mathcal{J}$:
\begin{align*}
    \boldsymbol{\ell}_j^t[a] = \dot{\boldsymbol{\ell}}^t[j, a] + \sum_{j' \in C_{j, a}}^{}\left\langle \boldsymbol{\ell}_{j'}^t, \boldsymbol{x}_{j'}^t\right\rangle.
\end{align*}
CFR then employs RM to minimize the total counterfactual regret $R_j^T = \max_{\boldsymbol{x}_{j}' \in \Delta^{n_j}}\sum_{t=1}^{T}\left\langle \boldsymbol{\ell}_j^t, \boldsymbol{x}_j^t - \boldsymbol{x}'_j\right\rangle $ at each decision node with respect to counterfactual loss.
It is guaranteed that the total regret $R^T$ is bounded by the sum of the total counterfactual regrets under each decision node, \ie, $R^T \le \sum_{j \in \mathcal{J}}^{}\left[{R}^T_j\right]^+$.

The computational procedure of CFR on each iteration $t$ is summarized as follows:
(1) decomposes the sequence-form strategy $\dot{\boldsymbol{x}}^t$ into local strategies $\boldsymbol{x}_j^t$ for each decision node $j\in \mathcal{J}$;
(2) recursively traverses the game tree to calculate the counterfactual loss $\boldsymbol{\ell}_j^t$;
(3) accumulates the \textbf{instantaneous counterfactual regret} $\boldsymbol{r}_j^t = \left\langle \boldsymbol{x}^t_{j}, \boldsymbol{\ell}^{t}_j\right\rangle \boldsymbol{1} - \boldsymbol{\ell}^t_j$ to obtain the \textbf{cumulative counterfactual regret} $\boldsymbol{R}_{j}^t = \boldsymbol{R}_j^{t-1} +  \boldsymbol{r}_{j}^t$;
(4) computes the new local strategy $\boldsymbol{x}_{j}^{t+1} = \left[\boldsymbol{R}_j^t\right]^+ / \left\Vert \left[\boldsymbol{R}_j^t\right]^+ \right\Vert_1$ for the next iteration;
(5) constructs $\dot{\boldsymbol{x}}^{t+1}$ based on local strategies;
(6) computes the \textbf{average strategy} $\bar{\boldsymbol{x}}^t = \boldsymbol{X}^t / \left\Vert \boldsymbol{X}^t \right\Vert_1$, where $\boldsymbol{X}^t = \boldsymbol{X}^{t-1} + \dot{\boldsymbol{x}}^t$ is the \textbf{cumulative strategy}.
If both players employ CFR, the average strategy profile $\left\langle \bar{\boldsymbol{x}}^T, \bar{\boldsymbol{y}}^T\right\rangle$ converges to a $\Delta(|\mathcal{J}_x| + |\mathcal{J}_y|)\sqrt{|\mathcal{A}|}/\sqrt{T}$-NE after $T$ iterations, where $\Delta$ is the range of losses.

\subsection{CFR Variants}
Since the birth of CFR, researchers have proposed many novel CFR variants, greatly improving the convergence rate of the vanilla CFR.
CFR+~\cite{tammelin_solving_2014,bowling_heads-up_2015} incorporates three small yet effective modifications, resulting in an order of magnitude faster convergence compared to CFR.
(1) CFR+ employs RM+ instead of RM, \ie, $\boldsymbol{R}^t_{j} = \left[\boldsymbol{R}^{t - 1}_{j} + \boldsymbol{r}^t_{j}\right]^+$.
(2) CFR+ adopts a linearly weighted average strategy where iteration $t$ is weighted by $t$, \ie, $\boldsymbol{X}^t =\boldsymbol{X}^{t-1} + t \dot{\boldsymbol{x}}^t$.
(3) CFR+ uses the alternating-updates technique.

DCFR~\cite{brown_solving_2019} is a family of algorithms which discounts prior iterations' cumulative regrets and dramatically accelerates convergence especially in games where some actions are very costly mistakes.
Specifically,
\resizebox{\linewidth}{!}{%
    \begin{minipage}{\linewidth}
        \begin{align*}
            \boldsymbol{R}^t_{j} = \boldsymbol{R}^{t - 1}_{j} \odot \boldsymbol{d}^{t-1}_{j} + \boldsymbol{r}^t_{j}, & \text{where}\;
            \boldsymbol{d}^t_{j}[a]  = \left\{
            \begin{array}{ll}
                \frac{t^\alpha}{t^\alpha + 1} & \text{if}\; \boldsymbol{R}^{t}_{j}[a]> 0 \\
                \frac{t^\beta}{t^\beta + 1}   & \text{otherwise}.                        \\
            \end{array}
            \right.                                                                                                                                                                 \\
            \boldsymbol{X}^t = \boldsymbol{X}^{t-1}                                                                  & \left(\frac{t-1}{t}\right)^\gamma  + \dot{\boldsymbol{x}}^t.
        \end{align*}
    \end{minipage}
}
Linear CFR~\cite{brown_solving_2019} is a special case of DCFR where iteration $t$'s contribution to cumulative regrets and cumulative strategy is proportional to $t$.

\subsection{The Predictive CFR Variant}
Recently, \cite{farina2021faster} demonstrates that a general regret minimization algorithm can be adapted for constructing a regret minimization algorithm tailored to a simplex. To illustrate, let us focus on the objective of minimizing the total counterfactual regret $R_j^T$ at a decision node $j$. Additionally, we have a general regret minimization algorithm $\mathbb{A}$, which we apply to a decision space $\mathbb{R}^{n_j}_{\ge 0}$.
At each iteration $t$, upon obtaining the counterfactual loss $\boldsymbol{\ell}_j^t$, we initially compute a modified loss $\tilde{\boldsymbol{\ell}}_j^t = - \boldsymbol{r}_j^t = \boldsymbol{\ell}_j^t - \left\langle \boldsymbol{\ell}_j^t, \boldsymbol{x}_j^t\right\rangle\boldsymbol{1}$. It is then fed into $\mathbb{A}$, and we receive its next decision $\tilde{\boldsymbol{x}}_j^{t+1} \in\mathbb{R}^{n_j}_{\ge 0}$. Finally, we obtain the next strategy as $\boldsymbol{x}_j^{t+1} = \tilde{\boldsymbol{x}}_j^{t+1} / \left\Vert \tilde{\boldsymbol{x}}_j^{t+1} \right\Vert_1$.

Based on the construction, \cite{farina2021faster} illustrates that employing OMD with $\psi=\frac{1}{2}\left\Vert \cdot \right\Vert_2^2$ as the algorithm $\mathbb{A}$ yields the same strategy as RM+ in each iteration. Additionally, by using Optimistic OMD with $\psi=\frac{1}{2}\left\Vert \cdot \right\Vert_2^2$ and $\boldsymbol{m}^{t+1} = -\boldsymbol{v}_j^{t+1}$, where $\boldsymbol{v}_j^{t+1} = \left\langle \boldsymbol{m}_j^{t+1}, \boldsymbol{x}_j^{t}\right\rangle \boldsymbol{1} - \boldsymbol{m}_j^{t+1}$ is the prediction of $\boldsymbol{r}_j^{t+1}$ and $\boldsymbol{m}_j^{t+1} = \boldsymbol{\ell}_j^{t}$ is the prediction of $\boldsymbol{\ell}_j^{t+1}$, Predictive RM+ (PRM+) is obtained, \ie,
\begin{align*}
    \resizebox{1\linewidth}{!}{$
        \boldsymbol{R}^{t}_j = [\boldsymbol{R}_j^{t-1} + \boldsymbol{r}_j^t] ^+, \tilde{\boldsymbol{R}}_j^{t+1}=[\boldsymbol{R}_j^t + \boldsymbol{v}_j^{t+1}]^+, \boldsymbol{x}_j^{t+1} = \tilde{\boldsymbol{R}}_j^{t+1} / \left\Vert \tilde{\boldsymbol{R}}_j^{t+1}\right\Vert_1.
    $}
\end{align*}
Moreover, \cite{farina2021faster} introduces Predictive CFR+ (PCFR+), which uses PRM+ as the regret minimization algorithm in each decision node and incorporates a quadratic weighted average strategy, defined as $\boldsymbol{X}^t =\boldsymbol{X}^{t-1} + t^2 \dot{\boldsymbol{x}}^t$.

We summarized CFR and typical variants in Table~\ref{Alg_Com}.

\section{Performance Decline of PCFR+ in the Presence of Dominated Actions}
\label{sec:motivation}

\begin{figure}[t]
    \centering
    \includegraphics[width=1\linewidth]{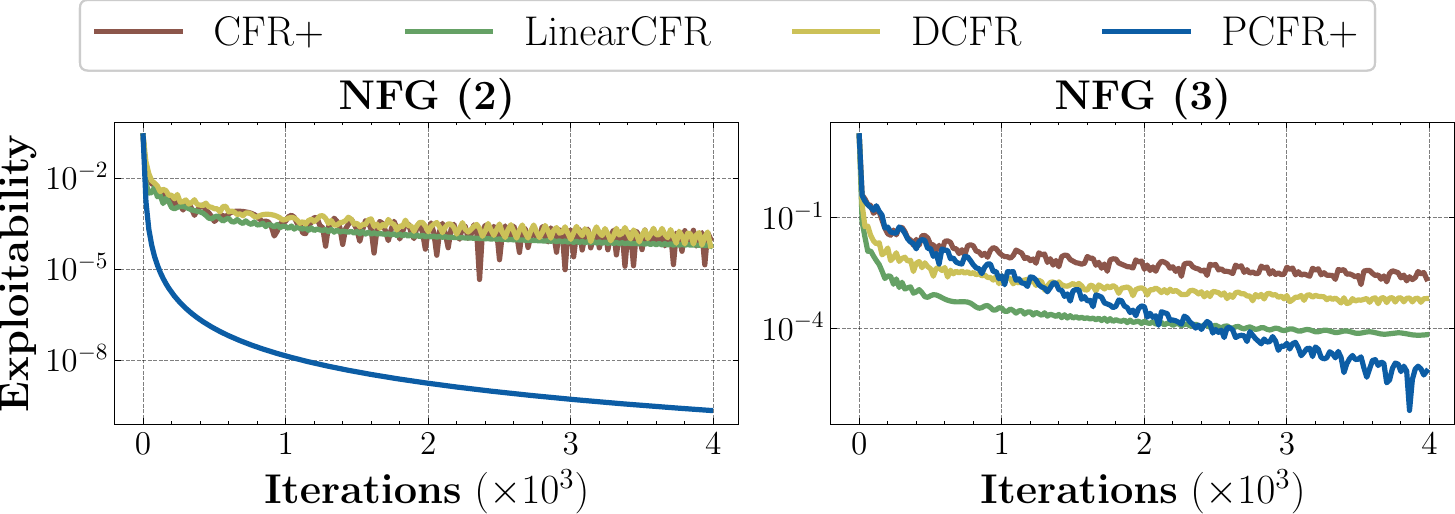}
    \caption{
        Convergence results of four CFR variants on two games.
    }
    \label{fig:motivation}
\end{figure}

Although PCFR+ successfully employs prediction to significantly accelerate convergence, we show that its performance deteriorates notably when facing dominated actions.
Consider a $2 \times 2$ zero-sum normal-form game \textbf{\textit{NFG (2)}}, represented by $\max_{\boldsymbol{x} \in \Delta^2}\min_{\boldsymbol{y} \in \Delta^2} \boldsymbol{x}^T\boldsymbol{U}\boldsymbol{y}$, where $\boldsymbol{U} = \left(\left(1,0\right),\left(0, 2\right)\right)$ is player 1's utility matrix.
As shown in the left plot of Figure~\ref{fig:motivation}, PCFR+ converges extremely faster than other CFR variants, benefiting from prediction during strategy updates.
However, when we introduce a dominated action for each player, resulting in \textbf{\textit{NFG (3)}}, where $\boldsymbol{U} = \left(\left(1,0, 5\right),\left(0, 2, 0\right), \left(0, 0, 100\right)\right)$, PCFR+'s performance declines greatly as depicted in the right plot.
For player 2, action 3 is dominated by action 1, allowing its elimination. Thus, player 1's action 3 can also be eliminated.
Consequently, \textit{NFG (3)} reduces to \textit{NFG (2)}, with the NE $\boldsymbol{x}=\boldsymbol{y}=\left(\frac{2}{3}, \frac{1}{3}, 0\right)$.
After the first iteration of PCFR+, player 2 has cumulative regrets $\boldsymbol{R}^1 = \left(\frac{100}{3}, \frac{100}{3}, 0\right)$.
The substantial loss of action 3 leads to high cumulative regrets for player 2's actions 1 and 2, making it challenging to distinguish between these two actions.
Due to the small instantaneous regrets per iteration, PCFR+ requires about 1500 iterations to achieve a 2:1 cumulative regrets for actions 1 and 2, producing an approximate NE.

While LinearCFR and DCFR initially converge faster than PCFR+ in \textit{NFG (3)} by assigning more weights to recent iterations and alleviating high regrets from earlier iterations, PCFR+ leverages predictions to surpass them after 2000 iterations.
It will be a best-of-both-worlds algorithm if we design a CFR variant that not only allocates more weights to recent iterations but also employs predictions to compute a new strategy, therefore greatly speeding up convergence.

\section{Minimizing Weighted Counterfactual Regret with OMD and Optimistic Variant}

In various situations, it makes sense to consider recent iterations as more crucial than earlier ones. Take equilibrium-finding algorithms as an example, earlier iterations are prone to selecting incorrect actions, resulting in substantial losses and causing significant regrets~\cite{brown_solving_2019}. Additionally, it is typical for recent iterations to generate strategies that are closer to a NE~\cite{perolat2021poincare}. Therefore, assigning more weights to recent iterations is a natural choice, potentially leading to fast convergence~\cite{abernethy2018faster,wang2018acceleration}.

Moreover, it is a prevalent practice to permit distinct weighting sequences for regrets and average strategies to facilitate fast convergence. For instance, CFR+ adopts a linearly weighted average strategy while employing a uniform weighting sequence for regrets. Besides, DCFR takes this approach a step further by using distinct weighting sequences for both regrets and average strategy.

In a weighted regret minimization algorithm applied to the sequence-form strategy space, the loss $\dot{\boldsymbol{\ell}}^t$ incurred at iteration $t$ is scaled by the weight $w_t$, forming a sequence of weights denoted as $\boldsymbol{w}$. The total weighted regret by the sequence $\boldsymbol{w}$ is
\begin{align*}
    R^T_{\boldsymbol{w}} = \max_{\dot{\boldsymbol{x}}' \in \mathcal{X}} \sum_{t=1}^{T} \left\langle w^t\dot{\boldsymbol{\ell}}^t, \dot{\boldsymbol{x}}^t - \dot{\boldsymbol{x}}'\right\rangle.
\end{align*}
Moreover, define $\bar{\boldsymbol{x}}^t_{\boldsymbol{\tau}}$ to be the weighted average strategy scaled by the weighting sequence $\boldsymbol{\tau}$:
\begin{align*}
    \bar{\boldsymbol{x}}^t_{\boldsymbol{\tau}} = \boldsymbol{X}^t_{\boldsymbol{\tau}} / \left\Vert \boldsymbol{X}^t_{\boldsymbol{\tau}} \right\Vert_1, \;\boldsymbol{X}^t_{\boldsymbol{\tau}} = \boldsymbol{X}_{\boldsymbol{\tau}}^{t-1} + \tau^t\dot{\boldsymbol{x}}^t.
\end{align*}

\begin{algorithm}[t]
    \KwIn{game $G$, total iterations $T$, general regret minimization algorithm $\mathbb{A}$, weighting sequences $\boldsymbol{w}$ and $\boldsymbol{\tau}$.}
    \For{$j \in \mathcal{J}$}{
    $\mathbb{A}_j \leftarrow $ instantiate $\mathbb{A}$ with $\mathcal{D} = \mathbb{R}^{n_j}_{\ge 0}$\;
    $\boldsymbol{x}_j^{1} \leftarrow \mathbb{A}_j$.first\_decision()\;
    }
    construct $\dot{\boldsymbol{x}}^{1}$ by $\left\{\boldsymbol{x}_j^{1}\mid j \in \mathcal{J} \right\}$\;
    \For{$t =1\rightarrow  T$}{
    decompose $\dot{\boldsymbol{x}}^t$ into $\left\{\boldsymbol{x}_j^t\mid j \in \mathcal{J} \right\}$\;
    calculate the counterfactual loss $\left\{\boldsymbol{\ell}_j^t \mid j \in \mathcal{J}\right\}$ based on the loss $\dot{\boldsymbol{\ell}}^t = A\dot{\boldsymbol{y}}^t$\;
    \For{$j \in \mathcal{J}$}{
    $\boldsymbol{r}_j^t \leftarrow   \left\langle \boldsymbol{\ell}_j^t, \boldsymbol{x}_j^t\right\rangle \boldsymbol{1} - \boldsymbol{\ell}_j^t$\;
    $\tilde{\boldsymbol{\ell}}_j^t \leftarrow   -w^t \boldsymbol{r}_j^t$\;
    $\mathbb{A}_j$.observe\_loss$(\tilde{\boldsymbol{\ell}}_j^t)$\;
    $\tilde{\boldsymbol{x}}_j^{t+1} \leftarrow$  $\mathbb{A}_j$.next\_decision()\;
    $\boldsymbol{x}_j^{t+1} \leftarrow   \tilde{\boldsymbol{x}}_j^{t+1} / \left\Vert \tilde{\boldsymbol{x}}_j^{t+1} \right\Vert_1$\;
    }
    construct $\dot{\boldsymbol{x}}^{t+1}$ by $\left\{\boldsymbol{x}_j^{t+1}\mid j \in \mathcal{J} \right\}$\;
    $\boldsymbol{X}^t \leftarrow \boldsymbol{X}^{t-1} + \tau^t\dot{\boldsymbol{x}}^t$\;
    $\bar{\boldsymbol{x}}^t \leftarrow \boldsymbol{X}^t / \left\Vert \boldsymbol{X}^t \right\Vert_1$\;
    }
    \KwOut{The final average strategy $\bar{\boldsymbol{x}}^T$.}
    \caption{Construction of a weighed CFR variant using a general regret minimization algorithm from player 1's perspective.}
    \label{algo}
\end{algorithm}

To minimize the total weighted regret $R^T_{\boldsymbol{w}}$, we first decompose it into the sum of the total weighted counterfactual regrets $R^T_{j, \boldsymbol{w}}$ under each decision node by constructing the counterfactual loss $\boldsymbol{\ell}_j^t$ at each iteration $t$, where
\begin{align*}
    R^T_{j, \boldsymbol{w}} = \max_{\boldsymbol{x}_{j}' \in \Delta^{n_j}}\sum_{t=1}^{T}\left\langle w^t\boldsymbol{\ell}_j^t, \boldsymbol{x}_j^t - \boldsymbol{x}'_j\right\rangle,
\end{align*}
so that we have the guarantee $R^T_{\boldsymbol{w}} \le \sum_{j \in \mathcal{J}}^{}[R_{j, \boldsymbol{w}}^T]^+$ and transform the objective to minimize the total weighted counterfactual regret $R^T_{j, \boldsymbol{w}}$ at each decision node.

To develop an algorithm for minimizing weighted counterfactual regret by leveraging a general regret minimization algorithm, we propose a construction method similar to the one described in~\cite{farina2021faster}, but with a modified loss $\tilde{\boldsymbol{\ell}}_j^t = -w^t\boldsymbol{r}_j^t$. A concise summary of this approach is presented in Algorithm~\ref{algo}.

As demonstrated in Theorem~\ref{thm:reduce} (all proofs of the theorems are in Appendix~\ref{app:proof}), when employing OMD with $\psi=\frac{1}{2} \left\Vert \cdot \right\Vert_2^2$ as the algorithm $\mathbb{A}$, it simplifies to the WCFR+ algorithm:
\begin{align*}
    \resizebox{1\linewidth}{!}{$
        \boldsymbol{R}^t_j = \left[\boldsymbol{R}_j^{t-1} + w^t \boldsymbol{r}_j^t\right]^+, \boldsymbol{x}_j^{t+1} = \left\Vert \boldsymbol{R}_j^t \right\Vert / \left\Vert \boldsymbol{R}_j^t \right\Vert_1, \boldsymbol{X}^t = \boldsymbol{X}^{t-1} + \tau^t\dot{\boldsymbol{x}}^t.
    $}
\end{align*}
Similarly, when employing optimistic OMD with $\psi=\frac{1}{2} \left\Vert \cdot \right\Vert_2^2$, it reduces to the PWCFR+ algorithm:
\begin{align*}
     & \boldsymbol{R}^t_j = \left[\boldsymbol{R}_j^{t-1} + w^t \boldsymbol{r}_j^t\right]^+, \tilde{\boldsymbol{R}}_j^{t+1} = \left[\boldsymbol{R}_j^t + w^{t+1} \boldsymbol{v}^{t+1}_j\right]^+,                        \\
     & \boldsymbol{x}_j^{t+1} = \left\Vert \tilde{\boldsymbol{R}}_j^{t+1} \right\Vert / \left\Vert \tilde{\boldsymbol{R}}_j^{t+1} \right\Vert_1, \boldsymbol{X}^t = \boldsymbol{X}^{t-1} + \tau^t\dot{\boldsymbol{x}}^t
\end{align*}
\begin{theorem}
    \label{thm:reduce}
    For all $\eta>0$, when employing OMD and optimistic OMD with $\psi=\frac{1}{2} \left\Vert \cdot \right\Vert_2^2$ as the algorithm $\mathbb{A}$, they reduce to WCFR+ and PWCFR+, respectively.
\end{theorem}

As illustrated in Theorem~\ref{thm:weighted_regret}, when both players employ a weighted regret minimization algorithm in a 2p0s IIG, the weighted average strategy profile converges to a NE.
\begin{theorem}
    \label{thm:weighted_regret}
    Assuming both players employ a weighted regret minimization algorithm with the weighting sequence $\boldsymbol{\tau}$ for the average strategy in a 2p0s IIG, and given that the two players have total weighted regrets $R^T_{ \boldsymbol{\tau}, x}$ and $R^T_{ \boldsymbol{\tau}, y}$ respectively, the weighted average strategy profile $(\bar{\boldsymbol{x}}_{\boldsymbol{\tau}}^T, \bar{\boldsymbol{y}}_{\boldsymbol{\tau}}^T)$ after $T$ iterations forms a $\frac{R^T_{\boldsymbol{\tau, x}} + R^T_{\boldsymbol{\tau}, y}}{\sum_{t=1}^{T}\tau^t}$-NE.
\end{theorem}

Although WCFR+ and PWCFR+ algorithms aim to minimize the total weight regret $R^T_{\boldsymbol{w}}$, we show that they also minimize the total weighted regret $R^T_{\boldsymbol{\tau}}$ as long as the weighting sequence $\boldsymbol{w}$ is more aggressive than $\boldsymbol{\tau}$.
\begin{theorem}
    \label{theorem-wcfr+}
    Assuming both players employ the WCFR+ algorithm with the weighting sequence $\boldsymbol{w}$ for loss and $\boldsymbol{\tau}$ for the average strategy in a 2p0s IIG, and $\{\frac{\tau_t}{w_t}\}_{t\le T}$ is a positive non-increasing sequence, the weighted average strategy profile $(\bar{\boldsymbol{x}}_{\boldsymbol{\tau}}^T, \bar{\boldsymbol{y}}_{\boldsymbol{\tau}}^T)$ after $T$ iterations forms a $\left(\sum_{j \in \mathcal{J}_x\cup \mathcal{J}_y}^{}\sqrt{\frac{\tau^1}{w^1}\sum_{t=1}^{T}\tau^t w^t\left\Vert \boldsymbol{r}_j^t\right\Vert_2^2  }\right)/ \sum_{t=1}^{T}\tau^t$-NE.
\end{theorem}
\begin{theorem}
    \label{theorem-pwcfr+}
    Assuming both players employ the PWCFR+ algorithm with the weighting sequence $\boldsymbol{w}$ for loss and $\boldsymbol{\tau}$ for the average strategy in a 2p0s IIG, and $\{\frac{\tau_t}{w_t}\}_{t\le T}$ is a positive non-increasing sequence, the weighted average strategy profile $(\bar{\boldsymbol{x}}_{\boldsymbol{\tau}}^T, \bar{\boldsymbol{y}}_{\boldsymbol{\tau}}^T)$ after $T$ iterations forms a $\left(\sum_{j \in \mathcal{J}_x\cup \mathcal{J}_y}^{}\!\sqrt{2\frac{\tau^1}{w^1}\sum_{t=1}^{T}\tau^t w^t\left\Vert \boldsymbol{r}_j^t\!-\boldsymbol{v}_j^t\right\Vert_2^2  }\right)/ \sum_{t=1}^{T}\tau^t$-NE.
\end{theorem}

A question remains regarding the selection of suitable weighting sequences in practical applications. Inspired by DCFR~\cite{brown_solving_2019} and to avoid potential numerical issues, we adopt a similar less-aggressive discounting sequence, resulting in a specific algorithm DCFR+:
\begin{align*}
     & \boldsymbol{R}_j^t = \left[\boldsymbol{R}_j^{t-1} \frac{(t-1)^{\alpha}}{(t-1)^{\alpha} + 1} + \boldsymbol{r}_j^t\right]^+, \boldsymbol{x}_j^{t+1} = \boldsymbol{R}_j^t / \left\Vert \boldsymbol{R}_j^t \right\Vert_1  , \\
     & \boldsymbol{X}^t = \boldsymbol{X}^{t-1} \left(\frac{t-1}{t}\right)^\gamma + \dot{\boldsymbol{x}}^t.
\end{align*}
DCFR+ integrates features from both CFR+ and DCFR in a principled manner.
It discounts cumulative regrets, assigning more weights to recent instantaneous regrets.
Additionally, it clips the negative part of cumulative regrets, allowing for quick reuse of promising actions.
Notably, the CFR variant discovered through evolutionary search~\cite{xu2022autocfr} emerges as a special case of DCFR+ where $\alpha=1.5$ and $\gamma=4$, showcasing faster convergence than DCFR.

We adopt a similar weighting sequence for PWCFR+, resulting in the PDCFR+ algorithm. PDCFR+ utilizes predicted cumulative regrets to compute new strategy akin to PCFR+, while updating cumulative regrets similar to DCFR+:
\begin{align*}
    \resizebox{1\linewidth}{!}{$
        \boldsymbol{R}_j^t = \left[\boldsymbol{R}_j^{t-1} \frac{(t-1)^{\alpha}}{(t-1)^{\alpha} + 1} + \boldsymbol{r}_j^t\right]^+,  \tilde{\boldsymbol{R}}_j^{t+1} = \left[\boldsymbol{R}_{j}^{t}\frac{t^\alpha}{t^\alpha + 1}+\boldsymbol{v}_j^{t+1}\right]^+,
    $}
\end{align*}
\begin{align*}
    \resizebox{1\linewidth}{!}{$
        \boldsymbol{x}_j^{t+1} = \tilde{\boldsymbol{R}}_j^{t+1} / \left\Vert \tilde{\boldsymbol{R}}_j^{t+1} \right\Vert_1  ,
        \boldsymbol{X}^t = \boldsymbol{X}^{t-1} \left(\frac{t-1}{t}\right)^\gamma + \dot{\boldsymbol{x}}^t.
    $}
\end{align*}

Interestingly, our connection between OMD and WCFR+ also provides a new perspective to understand the increasing weight $w^t$ in DCFR and DCFR+. When employing OMD with $\psi=\frac{1}{2}\left\Vert \cdot \right\Vert_2^2$ as the algorithm $\mathbb{A}$, it updates the decision $\tilde{\boldsymbol{x}}_j^{t+1}$, corresponding to the cumulative regrets $\boldsymbol{R}_j^{t}$ in WCFR+:
\begin{align*}
    \tilde{\boldsymbol{x}}_j^{t+1} = \underset{\tilde{\boldsymbol{x}}_j' \in \mathbb{R}_{\ge 0}^{n_j}}{\operatorname{argmin}}\;\left\{\left\langle -w^t\boldsymbol{r}^t_j, \tilde{\boldsymbol{x}}_j' \right\rangle + \frac{1}{2\eta}\left\Vert \tilde{\boldsymbol{x}}_j' - \tilde{\boldsymbol{x}}_j^{t} \right\Vert_2^2\right\}.
\end{align*}
It can be equivalently written as:
\begin{align*}
    \tilde{\boldsymbol{x}}_j^{t+1} & = \underset{\tilde{\boldsymbol{x}}_j' \in \mathbb{R}_{\ge 0}^{n_j}}{\operatorname{argmin}}\;\left\{\left\langle -\boldsymbol{r}^t_j, \tilde{\boldsymbol{x}}_j' \right\rangle + \frac{1}{2\eta w^t}\left\Vert \tilde{\boldsymbol{x}}_j' - \tilde{\boldsymbol{x}}_j^{t} \right\Vert_2^2\right\}             \\
                                   & = \underset{\tilde{\boldsymbol{x}}_j' \in \mathbb{R}_{\ge 0}^{n_j}}{\operatorname{argmin}}\;\left\{\left\langle -\boldsymbol{r}^t_j, \tilde{\boldsymbol{x}}_j' \right\rangle + \frac{1}{2\eta} \frac{1}{w^t} \left\Vert \tilde{\boldsymbol{x}}_j' - \tilde{\boldsymbol{x}}_j^{t} \right\Vert_2^2\right\}.
\end{align*}
We can interpret the increasing weight $w^t$ in two ways: an increasing learning rate $\eta^t = \eta w^t$ or a decreasing regularization term $\frac{1}{w_t}$. As a result, during the early iterations, cumulative regrets $\boldsymbol{R}_j^t$, \ie, $\tilde{\boldsymbol{x}}_j^{t+1}$, are learned at a gradual pace or regularized by a substantial strength. This interpretation aligns with our intuition, as we aim to control cumulative regrets from growing excessively large in the initial stages.

\section{Experiments}
In this section, we briefly describe testing games, compare PDCFR+ with other CFR variants, and analyze PDCFR+'s superior performance on normal-form games.
\subsection{Testing Games}

We use several commonly used IIGs in the research community and provide brief descriptions below. For more details, please refer to the Appendix~\ref{app:game}.
\textbf{\textit{Kuhn Poker}}~\cite{kuhn1950simplified} is a simplified poker with a three-card deck and one chance to bet for each player.
\textbf{\textit{Leduc Poker}}~\cite{southey2005bayes} is a larger game with a 6-card deck and two betting rounds.
In \textbf{\textit{Liar's Dice (x)}} ($x\small{=}4,5$)~\cite{lisy2015online}, each player gets an $x$-sided dice, which they roll at the start and take turn placing bets on the outcome.
\textbf{\textit{Goofspiel (x)}} ($x\small{=}4, 5$)~\cite{ross1971goofspiel} is a card game where each player has $x$ cards and aims to score points by bidding simultaneously in $x$ rounds.
\textbf{\textit{GoofspielImp (x)}} ($x\small{=}4, 5$) is a imperfect variant of \textit{Goofspiel (x)}, where bid cards remain unrevealed.
In \textbf{\textit{Battleship (x)}} ($x\small{=}2, 3$)~\cite{farina2019correlation}, players secretly position a $1 \times 2$ ship on separate $2 \times x$ grids and take turns firing at the opponent's ship three times.
\textbf{\textit{HUNL Subgame (x)}} ($x\small{=}3, 4$) is a heads-up no-limit Texas hold'em (HUNL) subgame generated by the top poker agent Libratus~\cite{brown2018superhuman}.

The testing games selected exhibit diverse structures and considerable game sizes, presenting a nontrivial challenge to solve.
However, their complexity remains manageable for an algorithm to complete within two weeks, rendering them highly suitable for evaluating algorithmic performance.
When extending to larger games, we can capitalize on a variety of well-established techniques, such as abstraction, decomposition~\cite{burch2014solving}, and subgame-solving~\cite{BrownS17,brown2018depth}. These approaches effectively mitigate the increased complexity associated with scaling up the size of the games.

\subsection{Convergence Results}
We compare PDCFR+ with CFR+, LinearCFR, DCFR, PCFR+ and DCFR+ across twelve games.
For DCFR, we adopt the hyperparameters suggested by the authors, specifically $\alpha=1.5$, $\beta=0$, and $\gamma=2$.
Regarding DCFR+ and PDCFR+, according to Theorems~\ref{theorem-wcfr+} and~\ref{theorem-pwcfr+}, $\alpha$ and $\gamma$ must ensure that $\{\frac{\tau_t}{w_t}\}_{t\le T}$ forms a positive, non-increasing sequence.
However, in practice, we find that the range of values for $\alpha$ and $\gamma$ can be larger, and they generally yield good convergence results.
Based on these observations, for DCFR+, we set $\alpha=1.5$ and $\gamma=4$ following~\cite{xu2022autocfr}, which is automatically set by an evolutionary search algorithm.
We perform a coarse grid search to fine-tune the hyperparameters for PDCFR+, and the best one, \ie, $\alpha=2.3$ and $\gamma=5$, is then used across all games.
All algorithms utilize the alternating-updates technique.

\begin{figure*}[t]
    \centering
    \includegraphics[width=1\linewidth]{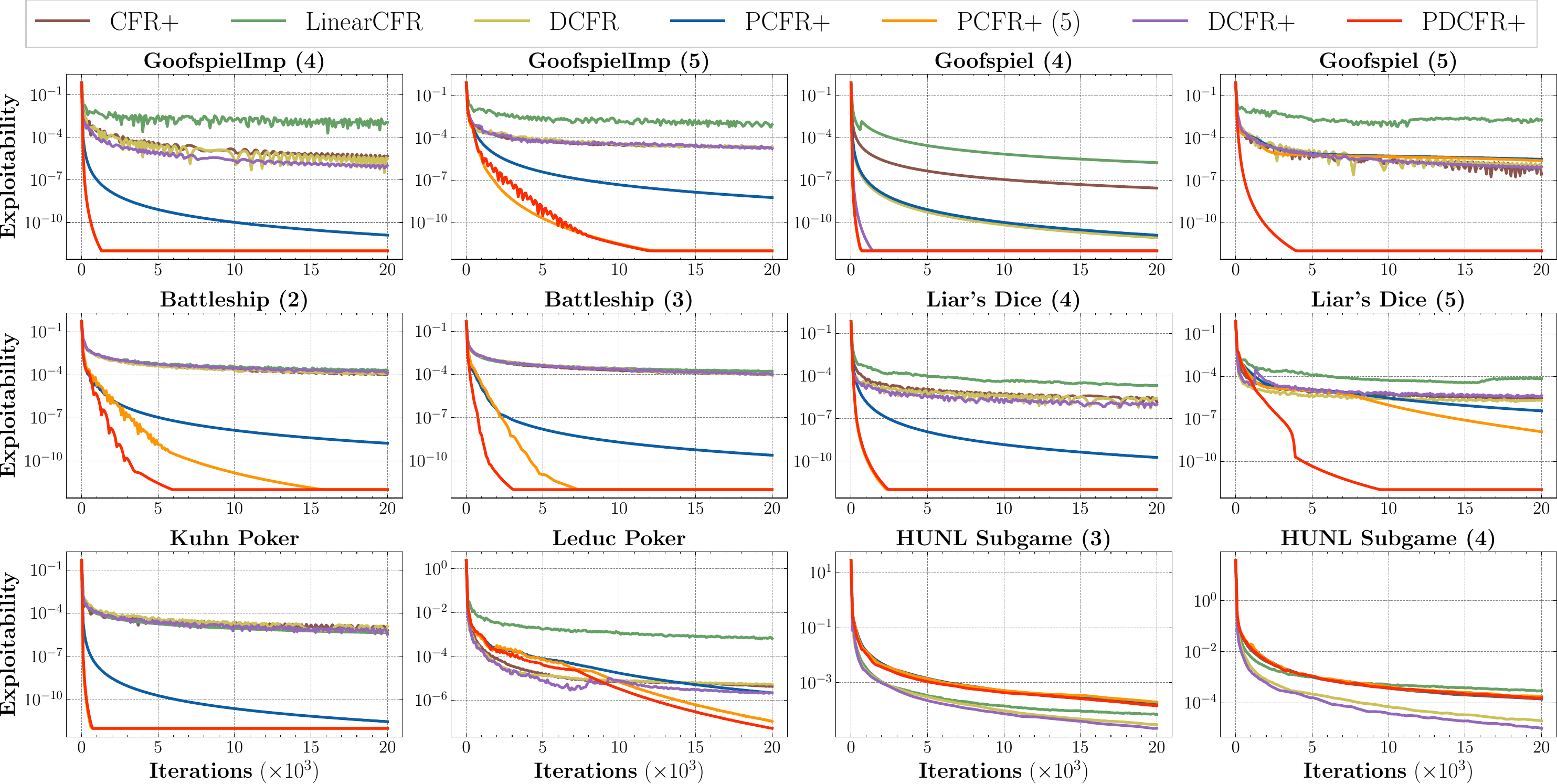}
    \caption{
        Convergence results of seven CFR variants on twelve testing games. Each algorithm runs for 20,000 iterations to display a long-time behavior. In all plots, the x-axis is the number of iteration, and the y-axis represents exploitability, displayed on a logarithmic scale.
    }
    \label{fig:compare}
\end{figure*}

\begin{figure}[t]
    \centering
    \includegraphics[width=1\linewidth]{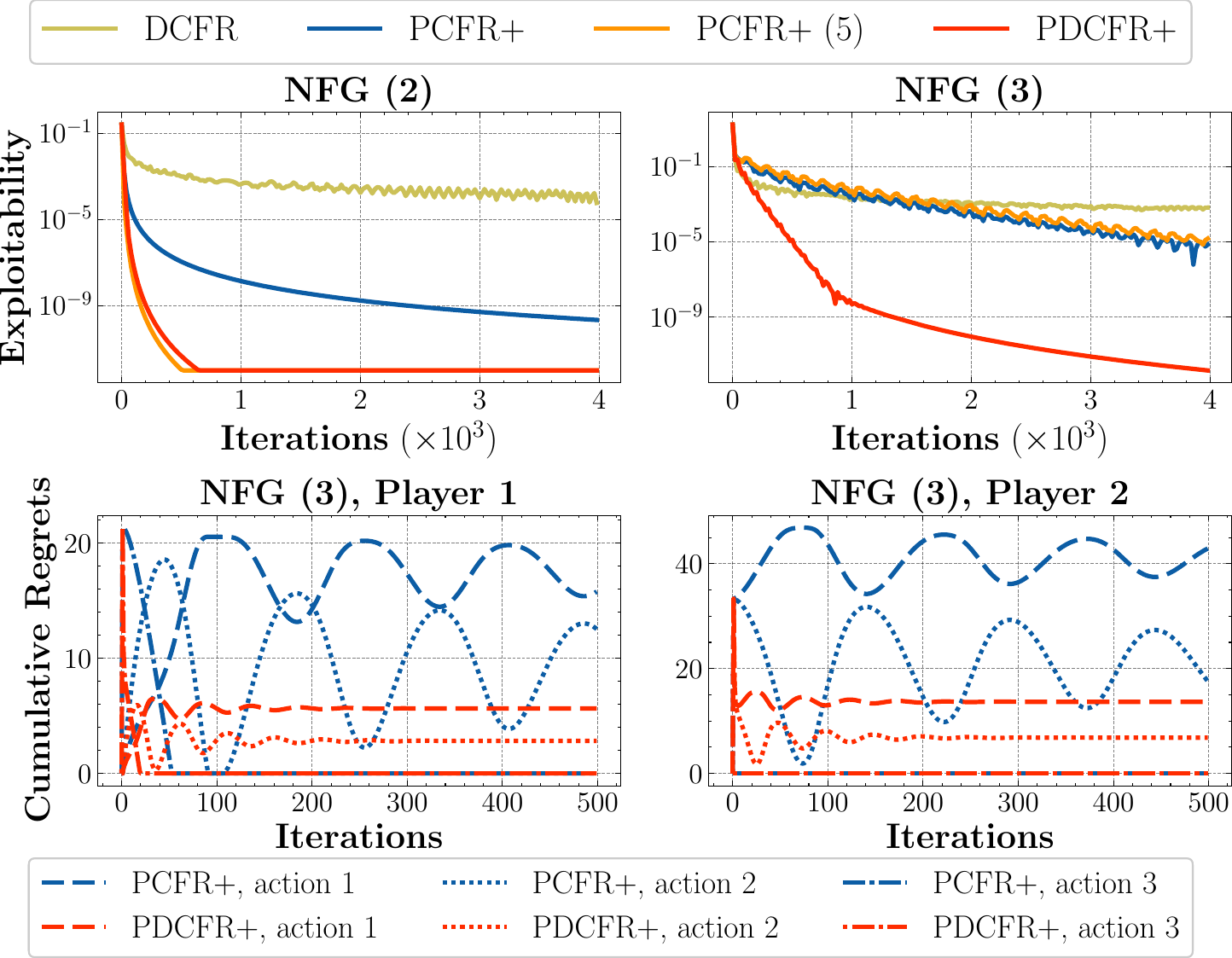}
    \caption{
        The top plots illustrate the convergence results of four CFR variants on two games.
        The bottom plots show that PDCFR+ quickly learns stable cumulative regrets in \textit{NFG (3)}.
    }
    \label{fig:analyse}
\end{figure}

We run each algorithm for 20,000 iterations in each testing game to observe their long-term behavior, with Figure~\ref{fig:compare} displaying exploitability curves.
We set a minimum reachable exploitability of $10^{-12}$, denoting the point where the average strategy profile is considered sufficiently converged to a NE.
Across non-poker games (\textit{Goofspiel}, \textit{Battleship}, \textit{Liar's Dice}), PDCFR+ outperforms others by 4-8 orders of magnitude, taking 2,000-12,000 iterations to achieve the minimum reachable exploitability. For poker games, PDCFR+ excels in \textit{Kuhn Poker}. In \textit{Leduc Poker}, PDCFR+ initially lags behind DCFR and CFR+ up to 10,000 iterations but exhibits faster convergence thereafter.
In the large-scale poker game \textit{HUNL Subgame}, PDCFR+ performs comparably to CFR+ and PCFR+, while DCFR+ emerges as the fastest algorithm.
In summary, PDCFR+ stands as an improved version of PCFR+, with a similar property for excellence in non-poker games.

PDCFR+ employs $\gamma=5$ in the computation of average strategies, while PCFR+ utilizes quadratic averaging, equivalent to $\gamma=2$ in PDCFR+. In order to assess whether PDCFR+'s improved performance is solely attributable to its specific averaging of strategies, we present the performance of PCFR+ (5), which employs $\gamma=5$, in Figure~\ref{fig:compare}. PCFR+ (5) outperforms PCFR+ with quadratic averaging in nine games.
PDCFR+ performs similarly to PCFR+ (5) in six games but exhibits significantly faster convergence in five games. Particularly in \textit{GoofSpiel (5)} and \textit{Liar's Dice (5)}, PDCFR+ surpasses PCFR+ (5) by 5-7 orders of magnitude. Thus, we conclude that the acceleration observed in PDCFR+ is not solely a result of using a more aggressive weighting scheme for determining average strategies but is also attributed to the weighting scheme employed in determining regrets.

To further comprehend the enhanced performance of PDCFR+, consider the normal-form games \textit{NFG (2)} and \textit{NFG (3)} discussed in section~\ref{sec:motivation}. Top plots of Figure~\ref{fig:analyse} show that PDCFR+ converges extremely fast on both games, not affected by the dominated actions.
After the first iteration of PDCFR+ in \textit{NFG (3)}, player 2 has high cumulative regrets $\boldsymbol{R}^1 = \left(\frac{100}{3}, \frac{100}{3}, 0\right)$ due to the dominated action 3.
As shown in the bottom-right plot of Figure~\ref{fig:analyse}, PDCFR+ rapidly discounts cumulative regrets, and at around the 200th iteration, it has already learned stable 2:1 cumulative regrets for actions 1 and 2, producing an approximate NE.
In contrast, PCFR+ is still struggling with fluctuations and requires much more iterations to approach a NE.
This highlights PDCFR+'s ability to swiftly mitigate the negative effects of dominated actions and consistently leverage predictions to accelerate convergence.

\section{Conclusions and Future Research}

This work proposes to minimize weighted counterfactual regret with OMD and optimistic variant. It provides  a new perspective to understand DCFR's superior performance and derives a novel CFR variant PDCFR+. PDCFR+ discounts cumulative regrets from early iterations to mitigate the negative effects of dominated actions and consistently leverages predictions to accelerate convergence. Experimental results demonstrate that PDCFR+ achieves competitive results compared with other CFR variants.
Recent work~\cite{farina2023regret} proposes two fixes for PRM+, achieving $O(1/T)$ convergence in normal-form games. It is worthwhile to investigate whether our algorithm can achieve similar convergence by introducing a specific weighting sequence.
Besides, combining PDCFR+ with function approximations~\cite{brown2019deep} and integrating it with the dynamic discounting framework~\cite{DDCFR} are also promising future works.

\section*{Acknowledgments}

This work is supported in part by the National Science and Technology Major Project (2022ZD0116401);
the Natural Science Foundation of China under Grant 62076238, Grant 62222606, and Grant 61902402;
the Jiangsu Key Research and Development Plan (No. BE2023016);
and the China Computer Federation (CCF)-Tencent Open Fund.

\bibliographystyle{named}
\bibliography{ref}

\newpage
\onecolumn
\appendix

\section{An illustration of Sequential Decision Process}
\label{app:sdp}

\begin{figure}[htbp]
    \centering
    \includegraphics[width=1\linewidth]{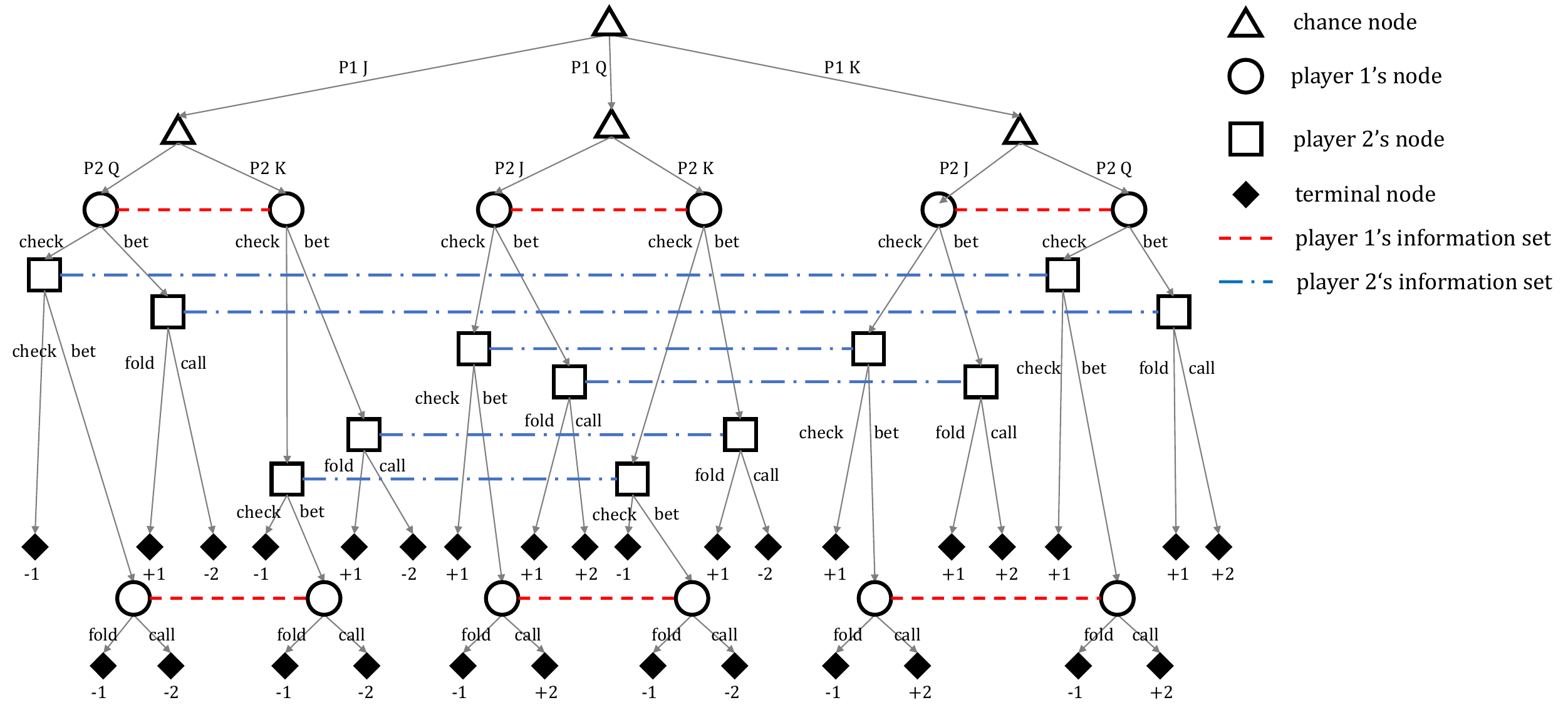}
    \caption{
        The complete game tree of \textit{Kuhn Poker}.
    }
    \label{fig:ext_game}
\end{figure}

To illustrate the sequential decision process, we use the game of \textit{Kuhn Poker} as an example. The typical formalism for describing an imperfect information game involves an extensive-form game forming a game tree as depicted in Figure~\ref{fig:ext_game}. The lack of information is represented by information sets for each player.

\begin{figure}[htbp]
    \centering
    \includegraphics[width=0.8\linewidth]{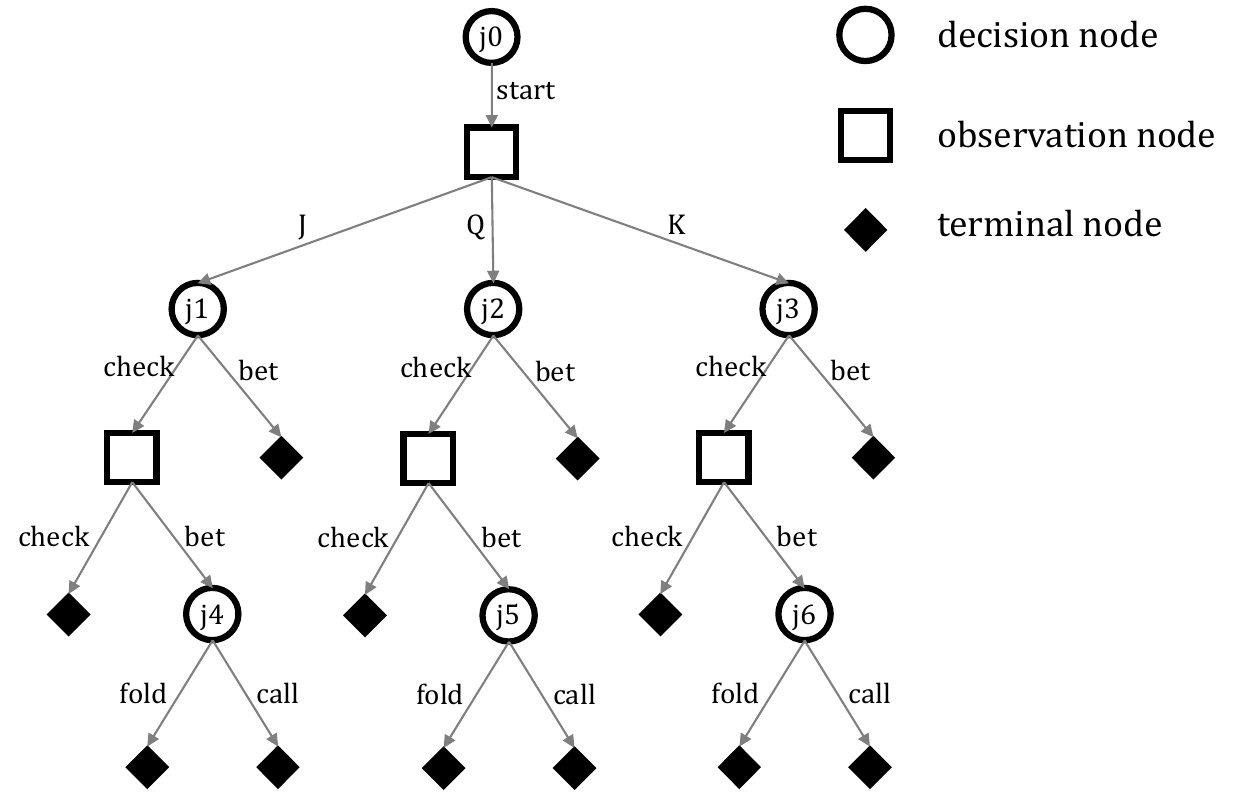}
    \caption{
        The sequential decision process for player 1 in the game of \textit{Kuhn Poker}.
    }
    \label{fig:sdp1}
\end{figure}

\begin{figure}[htbp]
    \centering
    \includegraphics[width=0.8\linewidth]{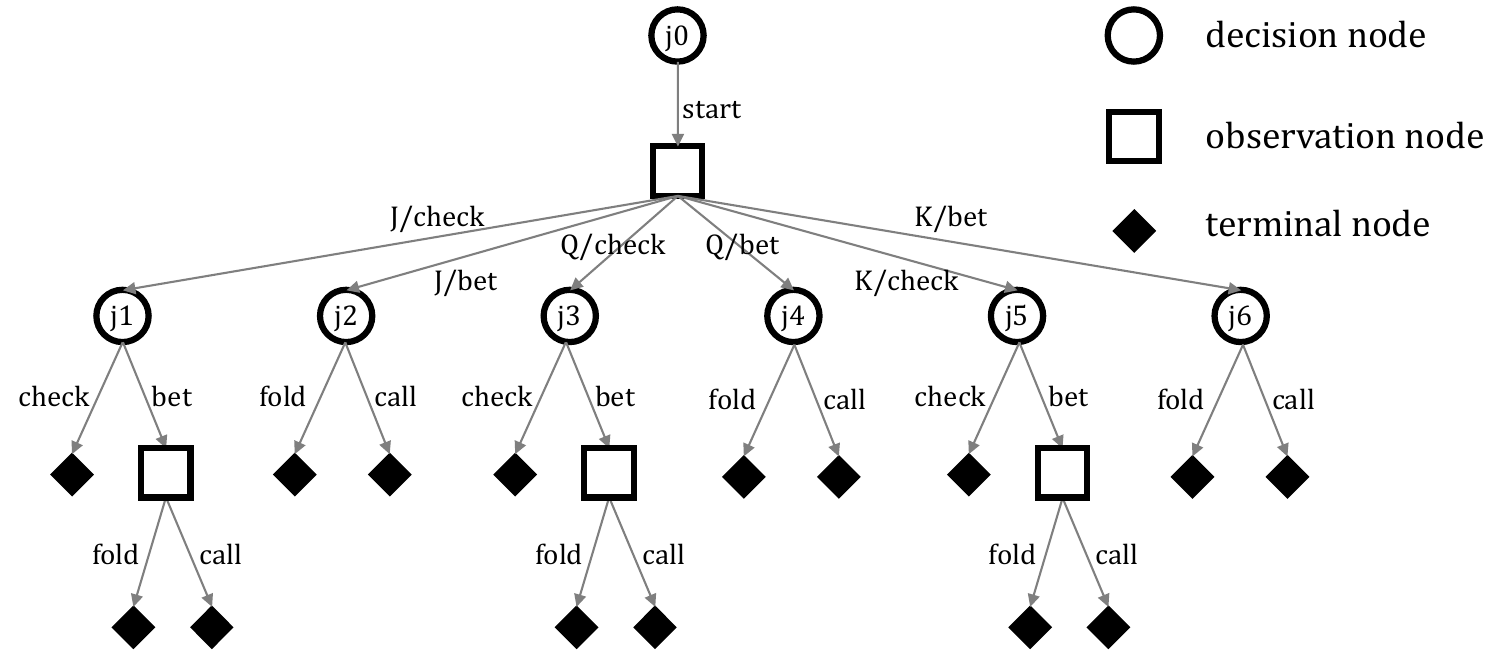}
    \caption{
        The sequential decision process for player 2 in the game of \textit{Kuhn Poker}.
    }
    \label{fig:sdp2}
\end{figure}

The sequential decision process encodes the decision problem confronted by an individual player. Figures~\ref{fig:sdp1} and ~\ref{fig:sdp2} showcase the sequential decision processes from the perspective of player 1 and 2, respectively. A one-to-one correspondence exists between information sets in the extensive-form game and decision nodes in the sequential decision process.

For player 1's decision process, there are seven decision nodes, \ie, $\mathcal{J}=\left\{j0, j1, j2, j3, j4, j5, j6\right\}$. At decision node $j0$, the action set is $\mathcal{A}_{j0}=\left\{start\right\}$ with earliest reachable decision nodes $C_{j0, start} = \left\{j1, j2, j3\right\}$ following the action $start$. Similarly, at decision node $j1$, the action set is $\mathcal{A}_{j1}= \left\{check, bet\right\}$ with earliest reachable decision nodes $C_{j1, check} = \left\{j4\right\}$ after taking the action $bet$.

There is a local strategy $\boldsymbol{x}_j \in \Delta^{n_j}$ for each decision node $j$. For example, $\boldsymbol{x}_{j1} \in \Delta^2$ is a vector indexed over $\left\{check, bet\right\}$ for decision node $j1$, where $\boldsymbol{x}_{j1}[check]$ and  $\boldsymbol{x}_{j2}[bet]$ represents the probabilities of selecting the actions $check$ and $bet$, respectively.
We can construct a sequence-form strategy $\dot{\boldsymbol{x}}$ based on local strategies.
For example, $\dot{\boldsymbol{x}}$ is vector indexed over $\{(j0, start), (j1, check), (j1, bet), (j2, check), (j2, bet), (j3, check), (j3, bet), (j4, fold), (j4, call), (j5, fold), (j5, call)$, $(j6, fold), (j6, call)\}$.
The probabilities in the sequence-form strategy are similar to the reach probabilities in the extensive-form game.
For instance, $\dot{\boldsymbol{x}}[(j1, bet)] = \boldsymbol{x}_{j0}[start]\boldsymbol{x}_{j1}[bet]$, and $\dot{\boldsymbol{x}}[(j6, fold)]=\boldsymbol{x}_{j0}[start]\boldsymbol{x}_{j3}[check]\boldsymbol{x}_{j6}[fold]$.

The payoff matrix $\boldsymbol{A}$ of the game \textit{Kuhn Poker} is presented in the Table~\ref{tab:payoff}. The first column of the matrix is player 1's sequence-form strategy, the first row of the matrix is player 2's sequence-form strategy. The matrix is sparse, and we only display the non-zero payoff for clarity. It is important to note that the payoff matrix $\boldsymbol{A}$ encodes the losses for player 1, while the utility for player 1 is listed in Figure~\ref{fig:ext_game}.

\begin{table}[htbp]
    \resizebox{1\linewidth}{!}{
        \begin{tabular}{|c|c|c|c|c|c|c|c|c|c|c|c|c|c|}
            \hline
                          & $(j0, start)$ & $(j1, check)$ & $(j1, bet)$ & $(j2, fold)$ & $(j2, call)$ & $(j3, check)$ & $(j3, bet)$ & $(j4, fold)$ & $(j4, call)$ & $(j5, check)$ & $(j5, bet)$ & $(j6, fold)$ & $(j6, call)$ \\ \hline
            $(j0,start)$  &               &               &             &              &              &               &             &              &              &               &             &              &              \\ \hline
            $(j1, check)$ &               &               &             &              &              & +1            &             &              &              & +1            &             &              &              \\ \hline
            $(j1, bet)$   &               &               &             &              &              &               &             & -1           & +2           &               &             & -1           & +2           \\ \hline
            $(j2, check)$ &               & -1            &             &              &              &               &             &              &              & +1            &             &              &              \\ \hline
            $(j2, bet)$   &               &               &             & -1           & -2           &               &             &              &              &               &             & -1           & +2           \\ \hline
            $(j3, check)$ &               & -1            &             &              &              & -1            &             &              &              &               &             &              &              \\ \hline
            $(j3, bet)$   &               &               &             & -1           & -2           &               &             & -1           & -2           &               &             &              &              \\ \hline
            $(j4, fold)$  &               &               &             &              &              &               & +1          &              &              &               & +1          &              &              \\ \hline
            $(j4, call)$  &               &               &             &              &              &               & +2          &              &              &               & +2          &              &              \\ \hline
            $(j5, fold)$  &               &               & +1          &              &              &               &             &              &              &               & +1          &              &              \\ \hline
            $(j5, call)$  &               &               & -2          &              &              &               &             &              &              &               & +2          &              &              \\ \hline
            $(j6, fold)$  &               &               & +1          &              &              &               & +1          &              &              &               &             &              &              \\ \hline
            $(j6, call)$  &               &               & -2          &              &              &               & -2          &              &              &               &             &              &              \\ \hline
        \end{tabular}
    }
    \caption{The payoff matrix of \textit{Kuhn Poker}.}
    \label{tab:payoff}
\end{table}

\section{Description of the Games}
\label{app:game}

\textbf{\textit{Kuhn Poker}} is a simplified form of poker proposed by Harold W. Kuhn~\cite{kuhn1950simplified}.
The game employs a deck of three cards, represented by J, Q, K.
At the beginning of the game, each player receives a private card drawn from a shuffled deck and places one chip into the pot.
The game involves four kinds of actions: 1) fold, giving up the current game, and the other player gets all the pot, 2) call, increasing his/her bet until both players have the same chips, 3) bet, putting more chips to the pot, and 4) check, declining to wager any chips when not facing a bet.
In \textit{Kuhn Poker}, each player has an opportunity to bet one chip.
If neither player folds, both players reveal their cards, and the player holding the higher card takes all the chips in the pot.
The utility for each player is defined as the difference between the number of chips after playing and the number of chips before playing.

\textbf{\textit{Leduc Poker}} is a larger poker game first introduced in~\cite{southey2005bayes}.
The game uses six cards that include two suites, each comprising three ranks (Js, Qs, Ks, Jh, Qh, Kh).
Similar to \textit{Kuhn Poker}, each player initially bets one chip, receives a single private card, and has the same set of action options.
In \textit{Leduc Poker}, the game unfolds over two betting rounds.
During the first round, players have an opportunity to bet two chips, followed by a chance to bet four chips in the second round.
After the first round, one public card is revealed.
If a player’s private card is paired with the public card, that player wins the game; otherwise, the player holding the highest private card wins the game.

\textbf{\textit{HUNL Subgame (x)}} ($x=3,4$) introduced in~\cite{brown_solving_2019} is a heads-up no-limit Texas hold'em(HUNL) sub-game generated by and solved in real-time by the state-of-the-art poker agent Libratus~\cite{brown2018superhuman}\footnote{\url{https://github.com/CMU-EM/LibratusEndgames}}.
In HUNL, both players (P1 and P2) start each hand with 20,000 chips, dealt two private cards from a standard 52-card deck.
P1 initially places 100 chips to the pot, followed by P2 adding 50 chips.
P2 starts the first round of betting.
Then players alternate in choosing to fold, call, check or raise.
A round ends when a player calls if both players have acted.
After the first round, three public cards are dealt face up for all players to observe, and P1 starts a similar round of betting.
In the third and fourth rounds, one additional public card is dealt and betting starts again with P1.
Unless a player has folded, the player with the best five-card poker hand, constructed from their two private cards and the five public cards, wins the pot.
In the case of a tie, the pot is split evenly.
\textit{HUNL Subgame (3)} begins at the start of the final betting round with 500 chips in the pot. \textit{HUNL Subgame (4)} begins at the start of the final betting round with 3,750 chips in the pot. In the first betting round, we use bet sizes of 0.5x, 1x the size of the pot, and an all-in bet. In other betting rounds, we use 1x the pot and all-in.

\textbf{\textit{Liar's Dice (x)}} ($x=4,5$)~\cite{lisy2015online} is a dice game where each player gets an $x$-sided dice and a concealment cup. At the beginning of the game, each player rolls their dice under their cup, inspecting the outcome privately. The first player then begins bidding of the form $p$-$q$, announcing that there are at least $p$ dices with the number of $q$ under all the cups. The highest dice number $x$ can be treated as any number. Then players take turns to take action: 1) bidding of the form $p$-$q$, $p$ or $q$ must be greater than the previous player's bidding, 2) calling `Liar', ending the game immediately and revealing all the dices. If the last bid is not satisfied, the player calling `Liar' wins the game. The winner's utility is 1 and the loser -1.

\textbf{\textit{Goofspiel (x)}} ($x=4,5$)~\cite{ross1971goofspiel} is a bidding card game. At the beginning of the game, each player receives $x$ cards numbered $1 \ldots x$, and there is a shuffled point card deck containing cards numbered $1 \ldots x$. The game proceeds in $x$ rounds. In each round, players select a card from their hand to make a sealed bid for the top revealed point card. When both players have chosen their cards, they show their cards simultaneously. The player who makes the highest bid wins the point card. If the bids are equal, the point card will be discarded. After $x$ rounds, the player with the most point cards wins the game. The winner's utility is 1 and the loser -1. We use a fixed deck of decreasing points.

\textbf{\textit{GoofspielImp (x)}} ($x=4,5$) is an imperfect information variant of \textit{Goofspiel (x)} where players are only told whether they have won or lost the bid, but not what the other player played.

\textbf{\textit{Battleship (x)}} ($x=2,3$)~\cite{farina2019correlation} is a classic board game where players secretly place a ship on their separate grids of size $2 \times x$ at the start of the game.
Each ship is $1 \times 2$ in size and has a value of 2.
Players take turns shooting at their opponent's ship, and the ship that has been hit at all its cells is considered sunk.
The game ends when one player's ship is sunk or when each player has completed three shots.
The utility for each player is calculated as the sum of the values of the opponent's sunk ship minus the sum of the values of their own lost ship.

We measure the sizes of the games in many dimensions and report the results in Table~\ref{tab:game}.
In the table, \emph{\#Histories} measures the number of histories in the game tree.
\emph{\#Infosets} measures the number of information sets in the game tree.
\emph{\#Terminal histories} measures the number of terminal histories in the game tree.
\emph{Depth} measures the depth of the game tree, \ie, the maximum number of actions in one history.
\emph{Max size of infosets} measures the maximum number of histories that belong to the same information set.

\begin{table}[htbp]
    \centering
    \caption{Sizes of the games.}
    \resizebox{1\linewidth}{!}{
        \begin{tabular}{c|rrrrr}
            \toprule
            Game             & \#Histories & \#Infosets & \#Terminal histories & Depth & Max size of infosets \\ \midrule
            Kuhn Poker       & 58          & 12         & 30                   & 6     & 2                    \\
            Leduc Poker      & 9,457       & 936        & 5,520                & 12    & 5                    \\
            Liar's Dice (4)  & 8,181       & 1,024      & 4,080                & 12    & 4                    \\
            Liar's Dice (5)  & 51,181      & 5,120      & 25,575               & 14    & 5                    \\
            Goofspiel (4)    & 1,077       & 270        & 576                  & 7     & 8                    \\
            Goofspiel (5)    & 26,931      & 3,252      & 14,400               & 9     & 48                   \\
            GoofspielImp (4) & 1,077       & 162        & 576                  & 7     & 14                   \\
            GoofspielImp (5) & 26,931      & 2,124      & 14,400               & 9     & 46                   \\
            Battleship (2)   & 10,069      & 3,286      & 5,568                & 9     & 4                    \\
            Battleship (3)   & 732,607     & 81,027     & 552,132              & 9     & 7                    \\
            HUNL Subgame (3) & 398,112,843 & 69,184     & 261,126,360          & 10    & 1,980                \\
            HUNL Subgame (4) & 244,005,483 & 43,240     & 158,388,120          & 8     & 1,980                \\
            \bottomrule
        \end{tabular}
    }
    \label{tab:game}
\end{table}

\section{Proof of Theorems}
\label{app:proof}
\subsection{Proof of Theorem~\ref{thm:reduce}}
\begin{proof}
    When employing OMD with $\psi=\frac{1}{2}\left\Vert \cdot \right\Vert_2^2$ as the algorithm $\mathbb{A}$, it updates the decision according to
    \begin{align*}
        \tilde{\boldsymbol{x}}_j^{t+1} & = \underset{\tilde{\boldsymbol{x}}_j'\in \mathbb{R}^n_{\ge 0}}{\operatorname{argmin}}\;\left\{\left\langle -w^t\boldsymbol{r}_j^t, \tilde{\boldsymbol{x}}_j'\right\rangle+\frac{1}{2\eta}\left\Vert \tilde{\boldsymbol{x}}_j'  - \tilde{\boldsymbol{x}}_j^t\right\Vert_2^2\right\}                                              \\
                                       & = \underset{\tilde{\boldsymbol{x}}_j'\in \mathbb{R}^n_{\ge 0}}{\operatorname{argmin}}\;\left\{\left\langle -2\eta w^t\boldsymbol{r}_j^t, \tilde{\boldsymbol{x}}_j'\right\rangle+\left\Vert \tilde{\boldsymbol{x}}_j'  - \tilde{\boldsymbol{x}}_j^t\right\Vert_2^2\right\}                                                       \\
                                       & = \underset{\tilde{\boldsymbol{x}}_j'\in \mathbb{R}^n_{\ge 0}}{\operatorname{argmin}}\;\left\{2\left\langle -\eta w^t\boldsymbol{r}_j^t, \tilde{\boldsymbol{x}}_j'\right\rangle+\left\Vert \tilde{\boldsymbol{x}}_j' \right\Vert_2^2 - 2\left\langle \tilde{\boldsymbol{x}}_j', \tilde{\boldsymbol{x}}_j^t\right\rangle\right\} \\
                                       & = \underset{\tilde{\boldsymbol{x}}_j' \in \mathbb{R}^n_{\ge 0}}{\operatorname{argmin}}\; \left\Vert \tilde{\boldsymbol{x}}_j' - \tilde{\boldsymbol{x}}_j^t  - \eta w^t \boldsymbol{r}_j^t \right\Vert_2^2                                                                                                                       \\
                                       & = [\tilde{\boldsymbol{x}}_j^t + \eta w^t \boldsymbol{r}_j^t]^+.
    \end{align*}
    The only effect of the step size $\eta$ is a rescaling of all decisions $\left\{\tilde{\boldsymbol{x}}_j^t\right\}$ by a constant. The output strategy $\boldsymbol{x}_j^{t+1} = \tilde{\boldsymbol{x}}_j^{t+1} / \left\Vert \tilde{\boldsymbol{x}}_j^{t+1} \right\Vert_1$ is invariant to positive rescaling of $\tilde{\boldsymbol{x}}_j^{t+1}$. So all choice of $\eta >0 $ result in the same strategy. Without loss of generality, we set $\eta = 1$ and $\boldsymbol{R}_j^t = \tilde{\boldsymbol{x}}_j^{t+1}$, which corresponds to the algorithm WCFR+.

    Similarly, when employing optimistic OMD with $\psi=\frac{1}{2}\left\Vert \cdot \right\Vert_2^2$ as the algorithm $\mathbb{A}$, it updates the decision according to
    \begin{align*}
        \tilde{\boldsymbol{z}}_j^{t} & = \underset{\tilde{\boldsymbol{z}}_j'\in \mathbb{R}^n_{\ge 0}}{\operatorname{argmin}}\;\left\{\left\langle -w^t\boldsymbol{r}_j^t, \tilde{\boldsymbol{z}}_j'\right\rangle+\frac{1}{2\eta}\left\Vert \tilde{\boldsymbol{z}}_j'  - \tilde{\boldsymbol{z}}_j^{t-1}\right\Vert_2^2\right\}                                              \\
                                     & = \underset{\tilde{\boldsymbol{z}}_j'\in \mathbb{R}^n_{\ge 0}}{\operatorname{argmin}}\;\left\{\left\langle -2\eta w^t\boldsymbol{r}_j^t, \tilde{\boldsymbol{z}}_j'\right\rangle+\left\Vert \tilde{\boldsymbol{z}}_j'  - \tilde{\boldsymbol{z}}_j^{t-1}\right\Vert_2^2\right\}                                                       \\
                                     & = \underset{\tilde{\boldsymbol{z}}_j'\in \mathbb{R}^n_{\ge 0}}{\operatorname{argmin}}\;\left\{2\left\langle -\eta w^t\boldsymbol{r}_j^t, \tilde{\boldsymbol{z}}_j'\right\rangle+\left\Vert \tilde{\boldsymbol{z}}_j' \right\Vert_2^2 - 2\left\langle \tilde{\boldsymbol{z}}_j', \tilde{\boldsymbol{z}}_j^{t-1}\right\rangle\right\} \\
                                     & = \underset{\tilde{\boldsymbol{z}}_j' \in \mathbb{R}^n_{\ge 0}}{\operatorname{argmin}}\; \left\Vert \tilde{\boldsymbol{z}}_j' - \tilde{\boldsymbol{z}}_j^{t-1}  - \eta w^t \boldsymbol{r}_j^t \right\Vert_2^2                                                                                                                       \\
                                     & = [\tilde{\boldsymbol{z}}_j^{t-1} + \eta w^t \boldsymbol{r}_j^t]^+,
    \end{align*}
    and
    \begin{align*}
        \tilde{\boldsymbol{x}}_j^{t+1} & = \underset{\tilde{\boldsymbol{x}}_j'\in \mathbb{R}^n_{\ge 0}}{\operatorname{argmin}}\;\left\{\left\langle -w^{t+1}\boldsymbol{v}_j^{t+1}, \tilde{\boldsymbol{x}}_j'\right\rangle+\frac{1}{2\eta}\left\Vert \tilde{\boldsymbol{x}}_j'  - \tilde{\boldsymbol{z}}_j^{t}\right\Vert_2^2\right\}                                              \\
                                       & = \underset{\tilde{\boldsymbol{x}}_j'\in \mathbb{R}^n_{\ge 0}}{\operatorname{argmin}}\;\left\{\left\langle -2\eta w^{t+1}\boldsymbol{v}_j^{t+1}, \tilde{\boldsymbol{x}}_j'\right\rangle+\left\Vert \tilde{\boldsymbol{x}}_j'  - \tilde{\boldsymbol{z}}_j^{t}\right\Vert_2^2\right\}                                                       \\
                                       & = \underset{\tilde{\boldsymbol{z}}_j'\in \mathbb{R}^n_{\ge 0}}{\operatorname{argmin}}\;\left\{2\left\langle -\eta w^{t+1}\boldsymbol{v}_j^{t+1}, \tilde{\boldsymbol{x}}_j'\right\rangle+\left\Vert \tilde{\boldsymbol{x}}_j' \right\Vert_2^2 - 2\left\langle \tilde{\boldsymbol{x}}_j', \tilde{\boldsymbol{z}}_j^{t}\right\rangle\right\} \\
                                       & = \underset{\tilde{\boldsymbol{x}}_j' \in \mathbb{R}^n_{\ge 0}}{\operatorname{argmin}}\; \left\Vert \tilde{\boldsymbol{x}}_j' - \tilde{\boldsymbol{z}}_j^{t}  - \eta w^{t+1} \boldsymbol{v}_j^{t+1} \right\Vert_2^2                                                                                                                       \\
                                       & = [\tilde{\boldsymbol{z}}_j^{t} + \eta w^{t+1} \boldsymbol{v}_j^{t+1}]^+,
    \end{align*}
    The only effect of the step size $\eta$ is a rescaling of all decisions $\left\{\tilde{\boldsymbol{x}}_j^t\right\}$ by a constant. The output strategy $\boldsymbol{x}_j^{t+1} = \tilde{\boldsymbol{x}}_j^{t+1} / \left\Vert \tilde{\boldsymbol{x}}_j^{t+1} \right\Vert_1$ is invariant to positive rescaling of $\tilde{\boldsymbol{x}}_j^{t+1}$. So all choice of $\eta >0 $ result in the same strategy. Without loss of generality, we set $\eta = 1$, $\boldsymbol{R}_j^t = \tilde{\boldsymbol{z}}_j^{t}$, and $\tilde{\boldsymbol{R}}_j^{t+1} = \tilde{\boldsymbol{x}}_j^{t+1}$, which corresponds to the algorithm PWCFR+.

\end{proof}

\subsection{Proof of Theorem~\ref{thm:weighted_regret}}

\begin{proof}
    In each iteration $t$, player 1 and player 2 generates a sequence-form strategy $\dot{\boldsymbol{x}}^t$ and $\dot{\boldsymbol{y}}^t$, respectively. The expected loss for player 1 is $\dot{\boldsymbol{x}}^\top \boldsymbol{A}\dot{\boldsymbol{y}}$. Subsequently, player 1 receives a loss vector $\dot{\boldsymbol{\ell}}_x^t = \boldsymbol{A}\dot{\boldsymbol{y}}^t$, while player 2 receives a loss vector $\dot{\boldsymbol{\ell}}_y^t = -\boldsymbol{A}^\top \dot{\boldsymbol{x}}^t$.

    For player 1's total weighted regret, we have
    \begin{align*}
        R^T_{\boldsymbol{\tau}, x} & = \max_{\dot{\boldsymbol{x}}' \in \mathcal{X}}\sum_{t=1}^{T} \left\langle \tau^t \dot{\boldsymbol{\ell}}_x^t, \dot{\boldsymbol{x}}^t - \dot{\boldsymbol{x}}
        '\right\rangle                                                                                                                                                                                                                                                                            \\
                                   & =\max_{\dot{\boldsymbol{x}}' \in \mathcal{X}}\sum_{t=1}^T \left\langle \tau^t \boldsymbol{A}\dot{\boldsymbol{y}}^t, \dot{\boldsymbol{x}}^t - \dot{\boldsymbol{x}}'\right\rangle                                                                              \\
                                   & = \sum_{t=1}^{T}\tau^t(\dot{\boldsymbol{x}}^t)^\top \boldsymbol{A}\dot{\boldsymbol{y}}^t - \min_{\dot{\boldsymbol{x}}' \in \mathcal{X}}\sum_{t=1}^{T}\tau^t \dot{\boldsymbol{x}}'^\top \boldsymbol{A}\dot{\boldsymbol{y}}^t                                  \\
                                   & = \sum_{t=1}^{T}\tau^t(\dot{\boldsymbol{x}}^t)^\top \boldsymbol{A}\dot{\boldsymbol{y}}^t - \min_{\dot{\boldsymbol{x}}' \in \mathcal{X}} \dot{\boldsymbol{x}}'^\top\boldsymbol{A} \left(
        \sum_{t=1}^{T}\tau^t\dot{\boldsymbol{y}}^t \right)                                                                                                                                                                                                                                        \\
                                   & = \sum_{t=1}^{T}\tau^t(\dot{\boldsymbol{x}}^t)^\top \boldsymbol{A}\dot{\boldsymbol{y}}^t -\left(\sum_{t=1}^{T}\tau^t\right)\min_{\dot{\boldsymbol{x}}' \in \mathcal{X}} \dot{\boldsymbol{x}}'^\top\boldsymbol{A} \bar{\boldsymbol{y}}_{\boldsymbol{\tau}}^T.
    \end{align*}
    Similarly, for player 2's total weighted regret, we have
    \begin{align*}
        R^T_{\boldsymbol{w}, y} & = \max_{\dot{\boldsymbol{y}}' \in \mathcal{Y}}\sum_{t=1}^{T} \left\langle \tau^t \boldsymbol{\ell}_y^t, \dot{\boldsymbol{y}}^t - \dot{\boldsymbol{y}}
        '\right\rangle                                                                                                                                                                                                                                                                            \\
                                & =\max_{\dot{\boldsymbol{x}}' \in \mathcal{X}}\sum_{t=1}^T \left\langle -\tau^t \boldsymbol{A}^\top\dot{\boldsymbol{x}}^t, \dot{\boldsymbol{y}}^t - \dot{\boldsymbol{y}}'\right\rangle                                                                           \\
                                & = -\sum_{t=1}^{T}\tau^t(\dot{\boldsymbol{x}}^t)^\top \boldsymbol{A}\dot{\boldsymbol{y}}^t + \max_{\dot{\boldsymbol{y}}' \in \mathcal{Y}}\sum_{t=1}^{T}\tau^t (\dot{\boldsymbol{x}}^t)^\top \boldsymbol{A}\dot{\boldsymbol{y}}'                                  \\
                                & = -\sum_{t=1}^{T}\tau^t(\dot{\boldsymbol{x}}^t)^\top \boldsymbol{A}\dot{\boldsymbol{y}}^t +\left(\sum_{t=1}^{T}\tau^t\right)\max_{\dot{\boldsymbol{y}}' \in \mathcal{Y}} (\bar{\boldsymbol{x}}_{\boldsymbol{\tau}}^T)^\top\boldsymbol{A} \dot{\boldsymbol{y}}'.
    \end{align*}
    So,
    \begin{align*}
         & \quad\delta_1(\bar{\boldsymbol{x}}_{\boldsymbol{\tau}}^T, \bar{\boldsymbol{y}}_{\boldsymbol{\tau}}^T) + \delta_2(\bar{\boldsymbol{x}}_{\boldsymbol{\tau}}^T, \bar{\boldsymbol{y}}_{\boldsymbol{\tau}}^T)                                                                                                                                                                                                                                                                \\
         & =(\bar{\boldsymbol{x}}_{\boldsymbol{w}}^T)^\top\boldsymbol{A}\bar{\boldsymbol{y}}_{\boldsymbol{w}}^T- \min_{\dot{\boldsymbol{x}}'\in \mathcal{X}}\dot{\boldsymbol{x}}'^\top\boldsymbol{A}\bar{\boldsymbol{y}}_{\boldsymbol{w}}^T + \max_{\dot{\boldsymbol{y}}' \in \mathcal{Y}}(\bar{\boldsymbol{x}}_{\boldsymbol{w}}^T)^\top \boldsymbol{A}\dot{\boldsymbol{y}}' - (\bar{\boldsymbol{x}}_{\boldsymbol{w}}^T)^\top\boldsymbol{A}\bar{\boldsymbol{y}}_{\boldsymbol{w}}^T \\
         & = \max_{\dot{\boldsymbol{y}}' \in \mathcal{Y}}(\bar{\boldsymbol{x}}_{\boldsymbol{w}}^T)^\top \boldsymbol{A}\dot{\boldsymbol{y}}' - \min_{\dot{\boldsymbol{x}}'\in \mathcal{X}}\dot{\boldsymbol{x}}'^\top\boldsymbol{A}\bar{\boldsymbol{y}}_{\boldsymbol{w}}^T                                                                                                                                                                                                           \\
         & = \frac{1}{\sum_{t=1}^{T}\tau^t}\left(R^T_{\boldsymbol{\tau}, x} + R^T_{\boldsymbol{\tau}, y}\right)                                                                                                                                                                                                                                                                                                                                                                    \\
    \end{align*}
    Hence, the weighted average strategy profile $(\bar{\boldsymbol{x}}_{\boldsymbol{\tau}}^T, \bar{\boldsymbol{y}}_{\boldsymbol{\tau}}^T)$ forms a $\frac{R^T_{\boldsymbol{\tau}, x}+R^T_{\boldsymbol{\tau}, y}}{\sum_{t=1}^{T}w^t}$-Nash equilibrium.
\end{proof}

\subsection{Proof of Theorem~\ref{theorem-wcfr+}}

The proof is based on~\cite[Theorem 6.8]{orabona2023modern}, with the addition of a weight to each iteration.

\begin{proof}
    According to the Theorem 2, we need to know the bound of $R^T_{\boldsymbol{\tau}}$.
    By decomposing the total weighted regret into the sum of the total weighted counterfactual regrets $R_{j, \boldsymbol{\tau}}^T$ under each decision node, we have $R^T_{\boldsymbol{\tau}} \le \sum_{j \in \mathcal{J}}^{}\left[R_{j, \boldsymbol{\tau}}^T\right]^+$.

    For the weighted counterfactual regret $R_{j, \boldsymbol{\tau}}^T$, we have
    \begin{align*}
        R_{j, \boldsymbol{\tau}}^{T} & = \max_{\boldsymbol{x}_j' \in \Delta^{n_j}}\sum_{t=1}^{T}\left\langle \tau^t\boldsymbol{\ell}_j^t, \boldsymbol{x}_j^t - \boldsymbol{x}_j'\right\rangle                                                                                                                            \\
                                     & = \max_{\boldsymbol{x}_j' \in \Delta^{n_j}}\sum_{t=1}^{T}\tau^t \left(\left\langle \boldsymbol{\ell}_j^t,  \boldsymbol{x}_j^t\right\rangle - \left\langle \boldsymbol{\ell}_j^t, \boldsymbol{x}_j'\right\rangle\right)                                                            \\
                                     & = \max_{\boldsymbol{x}_j' \in \Delta^{n_j}}\sum_{t=1}^{T}\tau^t \left(\left\langle \boldsymbol{\ell}_j^t,  \boldsymbol{x}_j^t\right\rangle\left\langle \boldsymbol{1}, \boldsymbol{x}_j'\right\rangle - \left\langle \boldsymbol{\ell}_j^t, \boldsymbol{x}_j'\right\rangle\right) \\
                                     & =\max_{\boldsymbol{x}_j' \in \Delta^{n_j}}\sum_{t=1}^{T}\tau^t \left\langle \left\langle \boldsymbol{\ell}_j^t,  \boldsymbol{x}_j^t\right\rangle\boldsymbol{1} - \boldsymbol{\ell}_j^t, \boldsymbol{x}_j'\right\rangle                                                            \\
                                     & = \max_{\boldsymbol{x}_j' \in \Delta^{n_j}}\sum_{t=1}^{T}\tau^t\left\langle  \boldsymbol{r}_j^t, \boldsymbol{x}_j'\right\rangle                                                                                                                                                   \\
                                     & = \max_{\boldsymbol{x}_j' \in \Delta^{n_j}}\sum_{t=1}^{T}\frac{\tau^t}{w^t} \left\langle w^t\boldsymbol{r}_j^t, \boldsymbol{x}_j'\right\rangle                                                                                                                                    \\
                                     & = \max_{\boldsymbol{x}_j' \in \Delta^{n_j}} \sum_{t=1}^{T}\frac{\tau^t}{w^t}\left\langle -\tilde{\boldsymbol{\ell}}_j^t, \boldsymbol{x}_j'\right\rangle
    \end{align*}

    When employing OMD with $\psi=\frac{1}{2}\left\Vert \cdot \right\Vert_2^2 $ as the algorithm $\mathbb{A}$, it updates the decision according to
    \begin{align*}
        \tilde{\boldsymbol{x}}_j^{t+1} = \underset{\tilde{\boldsymbol{x}}_j' \in \mathbb{R}^n_{\ge 0}}{\operatorname{argmin}}\; \left\{\left\langle -w^t \boldsymbol{r}_j^t, \tilde{\boldsymbol{x}}_j'\right\rangle + \frac{1}{2\eta}\left\Vert \tilde{\boldsymbol{x}}_j' - \tilde{\boldsymbol{x}}_j^t \right\Vert_2^2\right\}.
    \end{align*}
    $\forall \tilde{\boldsymbol{x}}_j' \in \Delta^{n_j}$, according to Lemma~\ref{lemma:orabona67}, we have that
    \begin{align*}
        \left\langle -w^t \boldsymbol{r}_j^t, \tilde{\boldsymbol{x}}_j^t - \tilde{\boldsymbol{x}}_j' \right\rangle & \le \frac{1}{\eta}\left(\frac{1}{2}\left\Vert \tilde{\boldsymbol{x}}'_j - \tilde{\boldsymbol{x}}_j^t \right\Vert_2^2 - \frac{1}{2}\left\Vert \tilde{\boldsymbol{x}}_j' - \tilde{\boldsymbol{x}}_j^{t+1} \right\Vert + \frac{\eta^2}{2}\left\Vert -w^t \boldsymbol{r}_j^t \right\Vert_2^2\right) \\
    \end{align*}
    Hence,
    \begin{align*}
         & \quad \sum_{t=1}^{T}\frac{\tau^t}{w^t}\left\langle -w^t \boldsymbol{r}_j^t, \tilde{\boldsymbol{x}}_j^t - \tilde{\boldsymbol{x}}_j' \right\rangle                                                                                                                                                                                                                                                                                                                                                                                          \\
         & \le \sum_{t=1}^{T}\frac{1}{\eta}\frac{\tau^t}{w^t}\left(\frac{1}{2}\left\Vert \tilde{\boldsymbol{x}}'_j - \tilde{\boldsymbol{x}}_j^t \right\Vert_2^2 - \frac{1}{2}\left\Vert \tilde{\boldsymbol{x}}_j' - \tilde{\boldsymbol{x}}_j^{t+1} \right\Vert + \frac{\eta^2}{2}\left\Vert -w^t \boldsymbol{r}_j^t \right\Vert_2^2\right)                                                                                                                                                                                                           \\
         & = \frac{1}{2\eta}\frac{\tau^1}{w^1}\left\Vert \tilde{\boldsymbol{x}}_j' - \tilde{\boldsymbol{x}}_j^1 \right\Vert_2^2  - \frac{1}{2\eta}\frac{\tau^T}{w^T} \left\Vert \tilde{\boldsymbol{x}}_j' - \tilde{\boldsymbol{x}}_j^{T+1} \right\Vert_2^2 + \frac{1}{2\eta}\sum_{t=1}^{T-1} \left(\frac{\tau^{t+1}}{w^{t+1}}  - \frac{\tau^{t}}{w^t}\right)\left\Vert \tilde{\boldsymbol{x}}_j' - \tilde{\boldsymbol{x}}_j^{t+1} \right\Vert_2^2  + \sum_{t=1}^{T}\frac{\eta}{2}\frac{\tau^t}{w^t}\left\Vert -w^t\boldsymbol{r}_j^t \right\Vert_2^2 \\
         & \le \frac{1}{2\eta}\frac{\tau^1}{w^1}+\sum_{t=1}^{T}\frac{\eta}{2}\frac{\tau^t}{w^t}\left\Vert  w^t\boldsymbol{r}_j^t\right\Vert_2^2                                                                                                                                                                                                                                                                                                                                                                                                      \\
    \end{align*}
    where we use $\left\{\frac{\tau^t}{w^t}\right\} $ a non-increasing sequence and $\tilde{\boldsymbol{x}}_j^1=\underset{\tilde{\boldsymbol{x}}'\in \mathbb{R}^{n_j}_{\ge 0}}{\operatorname{argmin}}\;\frac{1}{2}\left\Vert \tilde{\boldsymbol{x}}' \right\Vert_2^2 = \boldsymbol{0}$, $\tilde{\boldsymbol{x}}_j' \in \Delta^{n_j}$ in the last step.

    Using the fact that the strategies produced by WCFR+ do not depend on the chosen step size $\eta > 0$, we can choose the $\eta >0$ that minimizes the right hand side:
    \begin{align*}
        \sum_{t=1}^{T}\frac{\tau^t}{w^t}\left\langle -w^t \boldsymbol{r}_j^t, \tilde{\boldsymbol{x}}_j^t - \tilde{\boldsymbol{x}}_j'\right\rangle \le \sqrt{\frac{\tau^1}{w^1}\sum_{t=1}^{T}\tau^tw^t\left\Vert \boldsymbol{r}_j^t \right\Vert_2^2}
    \end{align*}
    Since WCFR+ chooses the next strategy in decision node $j$ as $\boldsymbol{x}_j^{t} = \tilde{\boldsymbol{x}}_j^t / \left\Vert \tilde{\boldsymbol{x}}^t_j \right\Vert_1$, we have
    \begin{align*}
        \left\langle \tilde{\boldsymbol{x}}_j^t, \boldsymbol{r}_j^t\right\rangle & = \left\langle \tilde{\boldsymbol{x}}_j^t, \left\langle \boldsymbol{\ell}_j^t, \boldsymbol{x}_j^t\right\rangle \boldsymbol{1} - \boldsymbol{\ell}_j^t \right\rangle                                                                                \\
                                                                                 & = \left\langle \boldsymbol{\ell}_j^t, \boldsymbol{x}_j^t \right\rangle\left\Vert \tilde{\boldsymbol{x}}_j^t \right\Vert_1  - \left\langle \tilde{\boldsymbol{x}}_j^t, \boldsymbol{\ell}_j^t\right\rangle                                           \\
                                                                                 & = \left\langle \boldsymbol{\ell}_j^t, \boldsymbol{x}_j^t\right\rangle\left\Vert \tilde{\boldsymbol{x}}_j^t \right\Vert_1 - \left\Vert \tilde{\boldsymbol{x}}_j^t \right\Vert_1\left\langle \boldsymbol{x}_j^t,  \boldsymbol{\ell}_j^t\right\rangle \\
                                                                                 & = 0
    \end{align*}
    So,
    \begin{align*}
        R_{j, \boldsymbol{\tau}}^T & = \max_{\tilde{\boldsymbol{x}}_j' \in \Delta^{n_j}} \sum_{t=1}^T \frac{\tau^t}{w^t}\left\langle w^t\boldsymbol{r}_j^t, \boldsymbol{x}_j'\right\rangle                                        \\
                                   & = \max_{\tilde{\boldsymbol{x}}_j' \in \Delta^{n_j}}\sum_{t=1}^{T}\frac{\tau^t}{w^t}\left\langle -w^t \boldsymbol{r}_j^t, \tilde{\boldsymbol{x}}_j^t - \tilde{\boldsymbol{x}}_j'\right\rangle \\
                                   & \le \sqrt{\frac{\tau^1}{w^1}\sum_{t=1}^{T}\tau^tw^t\left\Vert \boldsymbol{r}_j^t \right\Vert_2^2}
    \end{align*}
    and
    \begin{align*}
        R_{\boldsymbol{\tau}}^T \le \sum_{j \in \mathcal{J}}^{}\left[R_{j, \boldsymbol{\tau}}^T\right]^+ \le |\mathcal{J}|\sqrt{\frac{\tau^1}{w^1}\sum_{t=1}^{T}w^t\tau^t\left\Vert \boldsymbol{r}_j^t \right\Vert_2^2}
    \end{align*}
    Combining the Theorem 2, we have that the weighted average strategy profile $(\bar{\boldsymbol{x}}_{\boldsymbol{\tau}}^T, \bar{\boldsymbol{y}}_{\boldsymbol{\tau}}^T)$ after $T$ iterations forms a
    $\left(\sum_{j \in \mathcal{J}_x\cup \mathcal{J}_y}^{}\sqrt{\frac{\tau^1}{w^1}\sum_{t=1}^{T}\tau^t w^t\left\Vert \boldsymbol{r}_j^t\right\Vert_2^2  }\right)/ \left(\sum_{t=1}^{T}\tau^t\right)$
    -Nash equilibrium.
\end{proof}

\begin{lemma}{~\cite[Lemma 6.7]{orabona2023modern}}
    \label{lemma:orabona67}
    Let $\mathcal{D} \in \mathbb{R}^n$ be closed and convex, let $\boldsymbol{\ell} \in \mathbb{R}^n, \boldsymbol{x}^t \in \mathcal{D}$, and let $\psi:\mathcal{D} \rightarrow \mathbb{R}$ be a 1-strongly convex differentiable regularizer with respect to some norm $\left\Vert \cdot \right\Vert$, and let $\left\Vert \cdot \right\Vert_*$ be the dual norm to $\left\Vert \cdot \right\Vert$. Assume
    \begin{align*}
        \boldsymbol{x}^{t+1} := \underset{\boldsymbol{x}' \in \mathcal{D}}{\operatorname{argmin}}\; \left\{\left\langle \boldsymbol{\ell}^t, \boldsymbol{x}' \right\rangle + \frac{1}{\eta}\mathcal{B}_{\psi}(\boldsymbol{x}'\mid\mid \boldsymbol{x}^{t})\right\}
    \end{align*}
    Then $\forall \boldsymbol{x}' \in \mathcal{D}$, the following inequality holds:
    \begin{align*}
        \eta \left\langle \boldsymbol{\ell}^t, \boldsymbol{x}^t -\boldsymbol{x}' \right\rangle \le \mathcal{B}_{\psi}(\boldsymbol{x}'; \boldsymbol{x}^t) - \mathcal{B}_{\psi}(\boldsymbol{x}';\boldsymbol{x}^{t+1}) + \frac{\eta^2}{2}\left\Vert \boldsymbol{\ell} ^t\right\Vert^2_*
    \end{align*}
\end{lemma}

\subsection{Proof of Theorem~\ref{theorem-pwcfr+}}

The proof is based on~\cite[Proposition 5]{farina2021faster}, with the addition of a weight to each iteration.
\begin{proof}
    According to the Theorem 2, we need to know the bound of $R^T_{\boldsymbol{\tau}}$.
    By decomposing the total weighted regret into the sum of the total weighted counterfactual regrets $R_{j, \boldsymbol{\tau}}^T$ under each decision node, we have $R^T_{\boldsymbol{\tau}} \le \sum_{j \in \mathcal{J}}^{}\left[R_{j, \boldsymbol{\tau}}^T\right]^+$.

    For the weighted counterfactual regret $R_{j, \boldsymbol{\tau}}^T$, we have
    \begin{align*}
        R_{j, \boldsymbol{\tau}}^{T} & = \max_{\boldsymbol{x}_j' \in \Delta^{n_j}}\sum_{t=1}^{T}\left\langle \tau^t\boldsymbol{\ell}_j^t, \boldsymbol{x}_j^t - \boldsymbol{x}_j'\right\rangle                                                                                                                            \\
                                     & = \max_{\boldsymbol{x}_j' \in \Delta^{n_j}}\sum_{t=1}^{T}\tau^t \left(\left\langle \boldsymbol{\ell}_j^t,  \boldsymbol{x}_j^t\right\rangle - \left\langle \boldsymbol{\ell}_j^t, \boldsymbol{x}_j'\right\rangle\right)                                                            \\
                                     & = \max_{\boldsymbol{x}_j' \in \Delta^{n_j}}\sum_{t=1}^{T}\tau^t \left(\left\langle \boldsymbol{\ell}_j^t,  \boldsymbol{x}_j^t\right\rangle\left\langle \boldsymbol{1}, \boldsymbol{x}_j'\right\rangle - \left\langle \boldsymbol{\ell}_j^t, \boldsymbol{x}_j'\right\rangle\right) \\
                                     & =\max_{\boldsymbol{x}_j' \in \Delta^{n_j}}\sum_{t=1}^{T}\tau^t \left\langle \left\langle \boldsymbol{\ell}_j^t,  \boldsymbol{x}_j^t\right\rangle\boldsymbol{1} - \boldsymbol{\ell}_j^t, \boldsymbol{x}_j'\right\rangle                                                            \\
                                     & = \max_{\boldsymbol{x}_j' \in \Delta^{n_j}}\sum_{t=1}^{T}\tau^t\left\langle  \boldsymbol{r}_j^t, \boldsymbol{x}_j'\right\rangle                                                                                                                                                   \\
                                     & = \max_{\boldsymbol{x}_j' \in \Delta^{n_j}}\sum_{t=1}^{T}\frac{\tau^t}{w^t} \left\langle w^t\boldsymbol{r}_j^t, \boldsymbol{x}_j'\right\rangle                                                                                                                                    \\
                                     & = \max_{\boldsymbol{x}_j' \in \Delta^{n_j}} \sum_{t=1}^{T}\frac{\tau^t}{w^t}\left\langle -\tilde{\boldsymbol{\ell}}_j^t, \boldsymbol{x}_j'\right\rangle
    \end{align*}

    When employing optimistic OMD with $\psi=\frac{1}{2}\left\Vert \cdot \right\Vert_2^2 $ as the algorithm $\mathbb{A}$, it updates the decision according to
    \begin{align*}
        \tilde{\boldsymbol{z}}_j^{t} & = \underset{\tilde{\boldsymbol{z}}_j'\in \mathbb{R}^n_{\ge 0}}{\operatorname{argmin}}\;\left\{\left\langle -w^t\boldsymbol{r}_j^t, \tilde{\boldsymbol{z}}_j'\right\rangle+\frac{1}{2\eta}\left\Vert \tilde{\boldsymbol{z}}_j'  - \tilde{\boldsymbol{z}}_j^{t-1}\right\Vert_2^2\right\}
    \end{align*}
    \begin{align*}
        \tilde{\boldsymbol{x}}_j^{t+1} = \underset{\tilde{\boldsymbol{x}}_j' \in \mathbb{R}^n_{\ge 0}}{\operatorname{argmin}}\; \left\{\left\langle -w^{t+1} \boldsymbol{v}_{j}^{t+1}, \tilde{\boldsymbol{x}}_j'\right\rangle + \frac{1}{2\eta}\left\Vert \tilde{\boldsymbol{x}}_j' - \tilde{\boldsymbol{z}}_j^t \right\Vert_2^2\right\}.
    \end{align*}

    $\forall \tilde{\boldsymbol{x}}_j' \in \Delta^{n_j}$, we have
    \begin{align*}
        \sum_{t=1}^{T}\frac{\tau^t}{w^t}\left\langle -w^t \boldsymbol{r}_j^t, \tilde{\boldsymbol{x}}_j^t - \tilde{\boldsymbol{x}}_j' \right\rangle = \sum_{t=1}^{T}\frac{\tau^t}{w^t}\left(\left\langle -w^t\boldsymbol{r}_j^t + w^t\boldsymbol{v}_j^t, \tilde{\boldsymbol{x}}_j^t - \tilde{\boldsymbol{z}}_j^t\right\rangle + \left\langle -w^t\boldsymbol{v}_j^t, \tilde{\boldsymbol{x}}_j^t - \tilde{\boldsymbol{z}}_j^t\right\rangle + \left\langle -w^t \boldsymbol{r}_j^t, \tilde{\boldsymbol{z}}_j^t - \tilde{\boldsymbol{x}}_j'\right\rangle\right)
    \end{align*}
    According to Lemma~\ref{lemma:farina3} with $\rho=2\eta$,
    \begin{align*}
        \left\langle -w^t\boldsymbol{r}_j^t + w^t \boldsymbol{v}_j^t, \tilde{\boldsymbol{x}}_j^t - \tilde{\boldsymbol{z}}_j^t\right\rangle \le \eta\left\Vert w^t\boldsymbol{r}_j^t - w^t \boldsymbol{v}_j^t \right\Vert_2^2  + \frac{1}{4\eta} \left\Vert  \tilde{\boldsymbol{x}}_j^t - \tilde{\boldsymbol{z}}_j^t\right\Vert_2^2
    \end{align*}
    According to Lemma~\ref{lemma:farina4},
    \begin{align*}
        \left\langle -w^t\boldsymbol{v}_j^t, \tilde{\boldsymbol{x}}_j^t - \tilde{\boldsymbol{z}}_j^t\right\rangle \le\frac{1}{2\eta}\left(\left\Vert \tilde{\boldsymbol{z}}_j^t - \tilde{\boldsymbol{z}}_j^{t-1} \right\Vert_2^2 - \left\Vert \tilde{\boldsymbol{z}}_j^t - \tilde{\boldsymbol{x}}_j^t \right\Vert_2^2 - \left\Vert \tilde{\boldsymbol{x}}_j^t - \tilde{\boldsymbol{z}}_j^{t-1} \right\Vert_2^2\right)
    \end{align*}
    \begin{align*}
        \left\langle -w^t\boldsymbol{\ell}_j^t, \tilde{\boldsymbol{z}}_j^t - \tilde{\boldsymbol{x}}_j'\right\rangle \le\frac{1}{2\eta}\left(\left\Vert \tilde{\boldsymbol{x}}_j' - \tilde{\boldsymbol{z}}_j^{t-1} \right\Vert_2^2 - \left\Vert \tilde{\boldsymbol{x}}_j' - \tilde{\boldsymbol{z}}_j^t \right\Vert_2^2 - \left\Vert \tilde{\boldsymbol{z}}_j^t - \tilde{\boldsymbol{z}}_j^{t-1} \right\Vert_2^2\right)
    \end{align*}
    Hence, we have that for any $\boldsymbol{x}_j' \in \Delta^{n_j}$
    \begin{align*}
         & \quad\sum_{t=1}^{T}\frac{\tau^t}{w^t}\left\langle -w^t \boldsymbol{r}_j^t, \tilde{\boldsymbol{x}}_j^t - \tilde{\boldsymbol{x}}_j'\right\rangle                                                                                                                                                                                                                                                                                                                                                                                                                                                                           \\
         & \le \sum_{t=1}^{T}\frac{\tau^t}{w^t}\left(\eta\left\Vert w^t\boldsymbol{r}_j^t - w^t \boldsymbol{v}_j^t \right\Vert_2^2  + \frac{1}{4\eta} \left\Vert  \tilde{\boldsymbol{x}}_j^t - \tilde{\boldsymbol{z}}_j^t\right\Vert_2^2 + \frac{1}{2\eta}\left(\left\Vert \tilde{\boldsymbol{x}}_j' - \tilde{\boldsymbol{z}}_j^{t-1} \right\Vert_2^2 - \left\Vert \tilde{\boldsymbol{x}}_j' - \tilde{\boldsymbol{z}}_j^t \right\Vert_2^2 -\left\Vert\tilde{\boldsymbol{z}}_j^t - \tilde{\boldsymbol{x}}_j^{t}\right\Vert_2^2-\left\Vert \tilde{\boldsymbol{x}}_j^t - \tilde{\boldsymbol{z}}_j^{t-1} \right\Vert_2^2 \right)\right) \\
         & = \sum_{t=1}^{T}\frac{\tau^t}{w^t}\left(\eta \left\Vert w^t\boldsymbol{r}_j^t - w^t \boldsymbol{v}_j^t \right\Vert_2^2 - \frac{1}{4\eta}\left\Vert \tilde{\boldsymbol{x}}_j^t - \tilde{\boldsymbol{z}}_j^t \right\Vert_2^2  - \frac{1}{2\eta} \left\Vert \tilde{\boldsymbol{x}}_j^t - \tilde{\boldsymbol{z}}_j^{t-1}\right\Vert_2^2 + \frac{1}{2\eta}\left\Vert \tilde{\boldsymbol{x}}_j' - \tilde{\boldsymbol{z}}_j^{t-1} \right\Vert_2^2 - \frac{1}{2\eta}\left\Vert \tilde{\boldsymbol{x}}_j' - \tilde{\boldsymbol{z}}_j^{t} \right\Vert\right)                                                                       \\
         & \le \eta\sum_{t=1}^{T}\frac{\tau^t}{w^t}\left\Vert w^t\boldsymbol{r}_j^t - w^t \boldsymbol{v}_j^t\right\Vert_2^2 + \frac{1}{2\eta}\sum_{t=1}^{T}\frac{\tau^t}{w^t}\left(\left\Vert\tilde{\boldsymbol{x}}_j'   - \tilde{\boldsymbol{z}}_j^{t-1} \right\Vert_2^2 - \left\Vert\tilde{\boldsymbol{x}}_j' - \tilde{\boldsymbol{z}}_j^t\right\Vert_2^2\right)                                                                                                                                                                                                                                                                  \\
         & = \eta\sum_{t=1}^{T}\frac{\tau^t}{w^t}\left\Vert w^t\boldsymbol{r}_j^t- w^t \boldsymbol{v}_j^t\right\Vert_2^2 + \frac{1}{2\eta}\frac{\tau^1}{w^1}\left\Vert\tilde{\boldsymbol{x}}_j' - \tilde{\boldsymbol{z}}_j^{0}\right\Vert_2^2 - \frac{1}{2\eta}\frac{\tau^T}{w^T}\left\Vert\tilde{\boldsymbol{x}}_j' - \tilde{\boldsymbol{z}}_j^{T}\right\Vert_2^2 + \frac{1}{2\eta}\sum_{t=1}^{T-1}\left(\frac{\tau^{t+1}}{w^{t+1}} - \frac{\tau^{t}}{w^{t}}\right)\left\Vert\tilde{\boldsymbol{x}}_j' - \tilde{\boldsymbol{z}}_j^{t}\right\Vert_2^2                                                                               \\
         & \le \eta\sum_{t=1}^{T}\frac{\tau^t}{w^t}\left\Vert w^t\boldsymbol{r}_j^t- w^t \boldsymbol{v}_j^t\right\Vert_2^2 + \frac{1}{2\eta}\frac{\tau^1}{w^1}
    \end{align*}
    where we use $\left\{\frac{\tau^t}{w^t}\right\} $ a non-increasing sequence and $\tilde{\boldsymbol{z}}_j^0=\underset{\tilde{\boldsymbol{z}}'\in \mathbb{R}^{n_j}_{\ge 0}}{\operatorname{argmin}}\;\frac{1}{2}\left\Vert \tilde{\boldsymbol{z}}' \right\Vert_2^2 = \boldsymbol{0}$, $\tilde{\boldsymbol{x}}_j' \in \Delta^{n_j}$ in the last step.

    Using the fact that the strategies produced by PWCFR+ do not depend on the chosen step size $\eta > 0$, we can choose the $\eta >0$ that minimizes the right hand side:
    \begin{align*}
        \sum_{t=1}^{T}\frac{\tau^t}{w^t}\left\langle -w^t \boldsymbol{r}_j^t, \tilde{\boldsymbol{x}}_j^t - \tilde{\boldsymbol{x}}_j'\right\rangle \le \sqrt{2\frac{\tau^1}{w^1}\sum_{t=1}^{T}\tau^t w^t\left\Vert \boldsymbol{r}_j^t - \boldsymbol{v}_j^t\right\Vert_2^2  }
    \end{align*}

    Since PWCFR+ chooses the next strategy in decision node $j$ as $\boldsymbol{x}_j^{t} = \tilde{\boldsymbol{x}}_j^t / \left\Vert \tilde{\boldsymbol{x}}^t_j \right\Vert_1$, we have
    \begin{align*}
        \left\langle \tilde{\boldsymbol{x}}_j^t, \boldsymbol{r}_j^t\right\rangle & = \left\langle \tilde{\boldsymbol{x}}_j^t, \left\langle \boldsymbol{\ell}_j^t, \boldsymbol{x}_j^t\right\rangle \boldsymbol{1} - \boldsymbol{\ell}_j^t \right\rangle                                                                                \\
                                                                                 & = \left\langle \boldsymbol{\ell}_j^t, \boldsymbol{x}_j^t \right\rangle\left\Vert \tilde{\boldsymbol{x}}_j^t \right\Vert_1  - \left\langle \tilde{\boldsymbol{x}}_j^t, \boldsymbol{\ell}_j^t\right\rangle                                           \\
                                                                                 & = \left\langle \boldsymbol{\ell}_j^t, \boldsymbol{x}_j^t\right\rangle\left\Vert \tilde{\boldsymbol{x}}_j^t \right\Vert_1 - \left\Vert \tilde{\boldsymbol{x}}_j^t \right\Vert_1\left\langle \boldsymbol{x}_j^t,  \boldsymbol{\ell}_j^t\right\rangle \\
                                                                                 & = 0
    \end{align*}
    So,
    \begin{align*}
        R_{j, \boldsymbol{\tau}}^T & = \max_{\tilde{\boldsymbol{x}}_j' \in \Delta^{n_j}} \sum_{t=1}^T \frac{\tau^t}{w^t}\left\langle w^t\boldsymbol{r}_j^t, \boldsymbol{x}_j'\right\rangle                                        \\
                                   & = \max_{\tilde{\boldsymbol{x}}_j' \in \Delta^{n_j}}\sum_{t=1}^{T}\frac{\tau^t}{w^t}\left\langle -w^t \boldsymbol{r}_j^t, \tilde{\boldsymbol{x}}_j^t - \tilde{\boldsymbol{x}}_j'\right\rangle \\
                                   & \le \sqrt{2\frac{\tau^1}{w^1}\sum_{t=1}^{T}\tau^t w^t\left\Vert \boldsymbol{r}_j^t - \boldsymbol{v}_j^t\right\Vert_2^2  }
    \end{align*}
    and
    \begin{align*}
        R_{\boldsymbol{\tau}}^T \le \sum_{j \in \mathcal{J}}^{}\left[R_{j, \boldsymbol{\tau}}^T\right]^+ \le \sum_{j \in \mathcal{J}}^{}\sqrt{2\frac{\tau^1}{w^1}\sum_{t=1}^{T}\tau^t w^t\left\Vert \boldsymbol{r}_j^t - \boldsymbol{v}_j^t\right\Vert_2^2  }
    \end{align*}
    Combining the Theorem 2, we have that the weighted average strategy profile $(\bar{\boldsymbol{x}}_{\boldsymbol{\tau}}^T, \bar{\boldsymbol{y}}_{\boldsymbol{\tau}}^T)$ after $T$ iterations forms a
    $\left(\sum_{j \in \mathcal{J}_x\cup \mathcal{J}_y}^{}\sqrt{2\frac{\tau^1}{w^1}\sum_{t=1}^{T}\tau^t w^t\left\Vert \boldsymbol{r}_j^t - \boldsymbol{v}_j^t\right\Vert_2^2  }\right)/ \left(\sum_{t=1}^{T}\tau^t\right)$
    -Nash equilibrium.

\end{proof}

\begin{lemma}{~\cite[Lemma 3]{farina2021faster}}
    \label{lemma:farina3}
    For any $\boldsymbol{a}, \boldsymbol{b} \in \mathbb{R}^n$ and $\rho >0$, it holds that $\left\langle \boldsymbol{a}, \boldsymbol{b}\right\rangle \le \frac{\rho}{2}\left\Vert \boldsymbol{a} \right\Vert_*^2 + \frac{1}{2\rho} \left\Vert \boldsymbol{b} \right\Vert^2$.
\end{lemma}
\begin{lemma}{~\cite[Lemma 4]{farina2021faster}}
    \label{lemma:farina4}
    Let $\mathcal{D}\subseteq \mathbb{R}^n$ be closed and convex, let $\boldsymbol{\ell}^t \in \mathbb{R}^n, \boldsymbol{x}^t \in \mathcal{D}$, and let $\psi: \mathcal{D} \rightarrow  \mathbb{R}_{\ge 0}$ be a 1-strongly convex differentiable regularizer with respect to some norm $\left\Vert \cdot \right\Vert$, and let $\left\Vert \cdot \right\Vert_*$ be the dual norm to $\left\Vert \cdot \right\Vert$. Then,
    \begin{align*}
        \boldsymbol{x}^{t+1} := \underset{\boldsymbol{x}' \in \mathcal{D}}{\operatorname{argmin}}\; \left\{\left\langle \boldsymbol{\ell}^t, \boldsymbol{x}' \right\rangle + \frac{1}{\eta}\mathcal{B}_{\psi}(\boldsymbol{x}'\mid\mid \boldsymbol{x}^{t})\right\}
    \end{align*}
    is well defined (that is, the minimizer exists and is unique), and for all $\boldsymbol{x}' \in \mathcal{D}$ satisfies the inequality
    \begin{align*}
        \left\langle \boldsymbol{\ell}^t, \boldsymbol{x}^{t+1} - \boldsymbol{x}'\right\rangle \le \frac{1}{\eta}\left(\mathcal{B}_{\psi}(\boldsymbol{x}'\mid\mid \boldsymbol{x}^t) - \mathcal{B}_{\psi}(\boldsymbol{x}' \mid \mid \boldsymbol{x}^{t+1}) - \mathcal{B}_{\psi}(\boldsymbol{x}^{t+1} \mid\mid\boldsymbol{x}^t)\right)
    \end{align*}
\end{lemma}

\end{document}